\pdfoutput=1
\documentclass{article}
\usepackage{arxiv}

\makeatletter
\def\ps@pprintTitle{%
 \let\@oddhead\@empty
 \let\@evenhead\@empty
 \def\@oddfoot{}%
 \let\@evenfoot\@oddfoot}
\makeatother
\usepackage[utf8]{inputenc}
\usepackage[nospace]{cite}
\usepackage{lineno,hyperref}

\usepackage{graphicx}
\usepackage[acronym,toc,nonumberlist,nomain,nogroupskip]{glossaries}
\usepackage{glossary-mcols}
\setglossarystyle{mcolindex}
\usepackage{booktabs}
\usepackage{multirow}
\usepackage[table,xcdraw]{xcolor}
\usepackage{amsmath}
\usepackage{amsfonts}
\usepackage{bm}
\usepackage{soul}
\usepackage{enumitem}
\usepackage{arydshln}
\usepackage{subcaption}
\usepackage{comment}
\usepackage{geometry}
\usepackage[super]{nth}
\usepackage{pgfplots}
\pgfplotsset{compat=newest}
\usepgfplotslibrary{external}
\usepgfplotslibrary{patchplots}
\usetikzlibrary{shapes,arrows}
\usetikzlibrary{positioning,calc}
\usetikzlibrary{decorations.pathreplacing,decorations.markings,math,arrows.meta}
\usetikzlibrary{spy}
\usetikzlibrary{pgfplots.groupplots}
\usepgfplotslibrary{fillbetween}
\usetikzlibrary{patterns}
\newbox{\bigpicturebox}

\pgfdeclarelayer{ft}
\pgfdeclarelayer{bg}
\pgfsetlayers{bg,main,ft}

\graphicspath{{figures/}}
\usepackage{etoolbox}

\DeclareMathOperator*{\argmin}{arg\,min}
\usepackage[most]{tcolorbox}
\newtcolorbox{practicaltip}[1][]{%
  colback=yellow!12!white,
  colframe=yellow!75!orange,
  coltitle=black,
  boxrule=0.4pt, arc=2pt,
  fonttitle=\bfseries,
  title=Practical Tip, #1}
\newtcolorbox{shadednote}[1][]{%
  colback=cyan!7!white,
  colframe=cyan!35!blue!85!black,
  coltitle=white,
  boxrule=0.4pt, arc=2pt,
  fonttitle=\bfseries,
  title=Technical Insight, #1}
\newtcolorbox{workedexample}[1][]{%
  colback=black!3, colframe=black!50,
  boxrule=0.5pt, arc=3pt,
  fonttitle=\bfseries, title=Worked Example, #1}
\newtcolorbox{observation}[1][]{%
  colback=green!5, colframe=green!40!black,
  boxrule=0.4pt, arc=2pt,
  fonttitle=\bfseries, title=Key Finding, #1}
  
\newtcolorbox{figureplaceholder}[1][]{%
  colback=red!4, colframe=red!50!black,
  boxrule=0.6pt, arc=2pt,
  fonttitle=\bfseries\sffamily, title=Figure placeholder, #1}

\title{Deep Bayesian Filtering for Nonlinear Structural System Identification: A Tutorial on Structural and Physical Inductive Biases}

\author{
    Wei Liu$^{\star,\dagger}$\\
    National University of Singapore\\
    Singapore\\
    \texttt{weiliu@u.nus.edu}\\
    \And
    Konstantinos Vlachas$^{\star}$\\
    ETH Zurich\\
    Zurich, Switzerland\\
    \texttt{vlachas@ibk.baug.ethz.ch} 
    \And
    Eleni Chatzi$^{\dagger}$\\
    ETH Zurich\\
    Zurich, Switzerland\\
    \texttt{chatzi@ibk.baug.ethz.ch}
}

\makenoidxglossaries

\newacronym{DL}{DL}{Deep Learning}
\newacronym{DMM}{DMM}{Deep Markov Model}
\newacronym[longplural={Degrees of Freedom}]{DOF}{DOF}{Degree of Freedom}
\newacronym{EKF}{EKF}{Extended Kalman Filter}
\newacronym{ELBO}{ELBO}{Evidence Lower Bound}
\newacronym{FE}{FE}{Finite Element}
\newacronym{KF}{KF}{Kalman Filter}
\newacronym{KL}{KL}{Kullback-Leibler}
\newacronym{MAE}{MAE}{Mean Absolute Error}
\newacronym{NN}{NN}{Neural Network}
\newacronym{PgDMM}{PgDMM}{Physics-guided Deep Markov Model}
\newacronym{RMSE}{RMSE}{Root-Mean-Square Error}
\newacronym[longplural={Reduced-Order Models}]{ROM}{ROM}{Reduced-Order Model}
\newacronym{SHM}{SHM}{Structural Health Monitoring}
\newacronym{SINDy}{SINDy}{Sparse Identification of Nonlinear Dynamics}
\newacronym{VAE}{VAE}{Variational AutoEncoder}

\begin{document}
\maketitle

\def\thefootnote{*}\footnotetext{These authors contributed equally to this work\quad\textsuperscript{\textdagger}Corresponding authors}
\def\thefootnote{\arabic{footnote}}

\begin{abstract}
Nonlinear structural system identification remains challenging when governing physics is only partially known and measurements are sparse. Deep Bayesian filtering addresses these challenges by combining learnable dynamical representations with probabilistic state-space models, physical inductive biases, and uncertainty-aware inference. 
This paper presents, for the first time, a unified tutorial treatment of deep Bayesian filtering for nonlinear structural system identification, connecting model formulation, learning, posterior inference, and evaluation. Amortized variational inference and differentiable Kalman filtering are introduced as complementary strategies within this broader framework. The Deep Markov Model (DMM) provides a data-driven reference, while the Physics-guided Deep Markov Model (PgDMM) and Neural Extended Kalman Filter (Neural EKF) serve as illustrative instances through which methodological choices and their implications are examined. Dynamic state estimation and joint state and parameter estimation are demonstrated on a three-dimensional shear-frame benchmark with hysteretic Bouc--Wen connections and sparse acceleration measurements. The comparative analysis yields new observations concerning the relationships among reconstruction accuracy, predictive fidelity, physical interpretability, and estimation robustness. In these experiments, measurement-conditioned reconstruction errors underestimate open-loop forecasting errors by approximately three to five times. Physical structure improves latent-state interpretability without necessarily ensuring faithful predictive dynamics, while comparable typical reconstruction accuracy can conceal substantial differences in estimation robustness. These findings motivate practical guidelines for model construction, training, and evaluation that distinguish reconstruction, forecasting, state recovery, and parameter identification. The tutorial concludes by outlining challenges in uncertainty calibration, identifiability, and validation on real structures.
\end{abstract}

\keywords{system identification; deep Bayesian filtering; physics-guided learning; Deep Markov Models; Neural Kalman filtering}

\glsaddall
\glsunsetall
\printnoidxglossary[type=\acronymtype,title=Glossary of Acronyms]

\newcommand{\bA}{\bm{A}}
\newcommand{\bB}{\bm{B}}
\newcommand{\bC}{\bm{C}}
\newcommand{\bD}{\bm{D}}
\newcommand{\bE}{\bm{E}}
\newcommand{\bF}{\bm{F}}
\newcommand{\bG}{\bm{G}}
\newcommand{\bH}{\bm{H}}
\newcommand{\bI}{\bm{I}}
\newcommand{\bJ}{\bm{J}}
\newcommand{\bK}{\bm{K}}
\newcommand{\bL}{\bm{L}}
\newcommand{\bM}{\bm{M}}
\newcommand{\bN}{\bm{N}}
\newcommand{\bO}{\bm{O}}
\newcommand{\bP}{\bm{P}}
\newcommand{\bQ}{\bm{Q}}
\newcommand{\bR}{\bm{R}}
\newcommand{\bS}{\bm{S}}
\newcommand{\bT}{\bm{T}}
\newcommand{\bU}{\bm{U}}
\newcommand{\bV}{\bm{V}}
\newcommand{\bW}{\bm{W}}
\newcommand{\bX}{\bm{X}}
\newcommand{\bY}{\bm{Y}}
\newcommand{\bZ}{\bm{Z}}

\newcommand{\ba}{\bm{a}}
\newcommand{\bb}{\bm{b}}
\newcommand{\bc}{\bm{c}}
\newcommand{\bd}{\bm{d}}
\newcommand{\be}{\bm{e}}
\newcommand{\bff}{\bm{f}}
\newcommand{\bg}{\bm{g}}
\newcommand{\bh}{\bm{h}}
\newcommand{\bn}{\bm{n}}
\newcommand{\bp}{\bm{p}}
\newcommand{\bq}{\bm{q}}
\newcommand{\br}{\bm{r}}
\newcommand{\bs}{\bm{s}}
\newcommand{\bt}{\bm{t}}
\newcommand{\bu}{\bm{u}}
\newcommand{\bv}{\bm{v}}
\newcommand{\bw}{\bm{w}}
\newcommand{\bx}{\bm{x}}
\newcommand{\by}{\bm{y}}
\newcommand{\bz}{\bm{z}}

\newcommand{\balpha}{\bm{\alpha}}
\newcommand{\bbeta}{\bm{\beta}}
\newcommand{\bgamma}{\bm{\gamma}}
\newcommand{\bdelta}{\bm{\delta}}
\newcommand{\bepsilon}{\bm{\epsilon}}
\newcommand{\bzeta}{\bm{\zeta}}
\newcommand{\beeta}{\bm{\eta}}
\newcommand{\btheta}{\bm{\theta}}
\newcommand{\bkappa}{\bm{\kappa}}
\newcommand{\blambda}{\bm{\lambda}}
\newcommand{\bmu}{\bm{\mu}}
\newcommand{\bnu}{\bm{\nu}}
\newcommand{\bxi}{\bm{\xi}}
\newcommand{\brho}{\bm{\rho}}
\newcommand{\bsigma}{\bm{\sigma}}
\newcommand{\btau}{\bm{\tau}}
\newcommand{\bphi}{\bm{\phi}}
\newcommand{\bpsi}{\bm{\psi}}
\newcommand{\bomega}{\bm{\omega}}

\newcommand{\bGamma}{\bm{\Gamma}}
\newcommand{\bDelta}{\bm{\Delta}}
\newcommand{\bTheta}{\bm{\Theta}}
\newcommand{\bLambda}{\bm{\Lambda}}
\newcommand{\bXi}{\bm{\Xi}}
\newcommand{\bPi}{\bm{\Pi}}
\newcommand{\bSigma}{\bm{\Sigma}}
\newcommand{\bPhi}{\bm{\Phi}}
\newcommand{\bPsi}{\bm{\Psi}}
\newcommand{\bOmega}{\bm{\Omega}}

\newcommand\norm[1]{\left\lVert#1\right\rVert}

\newcommand\figureFontSize{9}

\section{Introduction}
System identification in structural dynamics encompasses inverse problems concerned with inferring unknown characteristics of dynamical systems from measured data.
Depending on the available observations and prior knowledge, these may include physical parameters, unmeasured states and responses, unknown inputs or excitations, and the governing model itself when its functional form is partially known or unavailable~\cite{quaranta2020review}.
Measurement noise, process variability, parameter uncertainty, and model-form uncertainty make uncertainty quantification an integral component of these tasks~\cite{fernandez2023bridge, Kamariotis2024}.
Timely and accurate identification informs condition assessment~\cite{black2021condition, reynders2014output}, decision-making~\cite{chiuso2019system}, and preventive maintenance~\cite{chen2023data, goutaudier2020impulse}, contributing to safer, more resilient, and more cost-effective infrastructure operation and management~\cite{argyroudis2022digital, iliopoulos2024remaining}.

This tutorial examines deep Bayesian filtering as a framework for learning partially known dynamics while retaining probabilistic inference and incorporating available physical knowledge.
Its organizing perspective is the progressive introduction of inductive biases into learnable probabilistic state-space models.
Starting from the \gls{DMM}, the discussion examines how assumptions about temporal evolution, latent-state structure, physical dynamics, observation mappings, and posterior inference shape the identification problem.
The \gls{PgDMM} and Neural \gls{EKF} serve as complementary examples of how these principles can be realized, rather than defining the scope of the framework.
The emphasis is on understanding what each modeling or inference choice contributes, how it is implemented, and what it does not guarantee.

Two identification settings illustrate this perspective.
In \emph{dynamic state estimation}, the physical system is fixed, while latent dynamics and observation mappings are learned to reconstruct, forecast, and interpret its response from sparse measurements.
In \emph{joint dynamic state and parameter estimation}, selected physical parameters are additionally treated as unknown quantities to be recovered alongside the latent state.
These settings address the identification of both unknown system quantities and unresolved components of the governing model, with physical structure incorporated where available.

System identification methods can be organized according to the amount and form of prior structure imposed on the dynamics.
Black-box approaches identify system characteristics without prescribing a detailed physical model~\cite{chen2018parameters, noel2017nonlinear, vlachas2025beyond, fernandez2024fault}.
They are used in fields such as finance~\cite{blach2011financial} and in operational modal analysis~\cite{gres2021kalman, volkmar2023experimental}, a central task of \gls{SHM}~\cite{au2017operational}.
Their flexibility nevertheless has limitations: some formulations assume linearity or time invariance, while more expressive models may offer limited interpretability and poor robustness under extrapolation~\cite{thelen2022comprehensive}.

Although linear identification has well-established theoretical foundations~\cite{caines2018linear}, engineering systems can exhibit nonlinear and nonstationary behavior.
Examples include structural systems undergoing yielding under strong seismic excitation~\cite{cattari2022nonlinear, moya2025post}, aerospace structures whose mechanical joints exhibit friction and hysteresis~\cite{cong2024hysteretic, tan2022study, yang2026unified, vlachas2025reduced}, and wind-turbine blades or hull panels experiencing large deformations under stochastic loading~\cite{cross2014nonlinear, fernandez2025gaussian, moynihan2023system}.
Identification must therefore yield models that capture observed responses and remain physically meaningful and reliable beyond the measured data.
Applications such as virtual sensing and decision support additionally depend on model fidelity and may impose regulatory requirements and real-time computational constraints~\cite{liang2024harnessing}.
These demands motivate methods that combine representational flexibility, physical interpretability, explicit uncertainty quantification, and reliable prediction.

Hybrid or gray-box identification incorporates physical knowledge while learning model components that describe unresolved nonlinearities, discrepancies, or mechanisms.
Classical formulations typically prescribe a finite-dimensional model structure, control its complexity through model order~\cite{billings2013nonlinear, worden2019nonlinearity}, and estimate its parameters from measured data~\cite{cheng2023machine, avendano2025sequential}.
Related approaches to nonlinear identification include time-domain nonlinear subspace identification~\cite{liu2018modified, noel2014subspace, yao2025stiffness}, nonlinear autoregressive moving-average models with exogenous inputs~\cite{chen2007genetic, dong2023neural}, signal-analysis methods based on wavelet, short-time Fourier, and Hilbert transforms~\cite{feldman2011hilbert, heller2009experimental, lichota2023wavelet}, and Gaussian-process models~\cite{avendano2017gaussian, fuentes2021equation, ramirez2018non}.

Model-discovery approaches extend identification to the governing equations themselves.
Symbolic regression constructs explicit mathematical expressions from specified operators and functions~\cite{makke2024interpretable}, using search strategies to balance agreement with observations and expression complexity~\cite{kamienny2022end}.
Genetic programming evolves candidate expressions within potentially vast search spaces, whereas grammar-based approaches constrain their construction through prescribed rules~\cite{angelis2023artificial, worden2018evolutionary, yu2025grammar}.
Dictionary-based methods provide a related route, with \gls{SINDy} as a prominent example~\cite{brunton2016discovering, naozuka2022sindy}.
Here, the dynamics are assumed to admit a sparse representation in a predefined function library, and identification selects active terms and estimates their coefficients.
Success depends on the suitability of the library and the reliability of sparse selection, motivating several extensions for system identification~\cite{fasel2022ensemble, kaiser2018sparse, lai2021finding, viknesh2024adam, voina2024deep}.

Such flexibility can entail substantial computational and data requirements, careful library design, or heuristic search procedures with limited theoretical guarantees~\cite{ljung2020deep, polson2017deep}.
Moreover, expressions selected primarily for agreement with observations may violate known physical constraints, limiting their interpretation and reliability in subsequent engineering applications~\cite{cheng2023machine, ghadami2022data}.
Other methods instead take governing equations as a starting point and infer unknown quantities by enforcing consistency with both physics and measurements.
Physics-informed neural networks are a prominent example, although scalability, generalization, and computational efficiency remain challenging~\cite{haywood2025response, jian2025modal, moradi2023novel, zhang2024dual}.
Optimizing a Discrete Loss (ODIL) directly optimizes discrete solution variables using equation residuals and observation errors, avoiding a neural-network representation of the solution and achieving substantial computational savings in several reported applications~\cite{amoudruz2025bayesian, karnakov2024solving}.
These methods are attractive when sufficiently informative governing models are available and typically formulate estimation as optimization over a dataset or observation window.

For continuously monitored systems, inference must also accommodate sequential measurements.
Observer formulations combine a dynamical model with incoming observations to recursively update estimates of unmeasured states and, where included, unknown inputs, parameters, or responses~\cite{bernard2019observer, boutat2021observer, niven2024dynamical}.
Once specified or learned, the model can support online estimation for monitoring, virtual sensing, diagnostics, and decision support.
Observers may employ conventional nonlinear models, learned representations, or combinations of the two~\cite{dikshit2025scalable, gedon2021deep, jaeger2002adaptive}.

Bayesian filtering provides a probabilistic framework for sequential inference and uncertainty propagation~\cite{koune2023bayesian, yan2025bayesian, zhou2022sparse}.
Its state-space formulation comprises two principal components~\cite{chatzi2009unscented}:
\begin{enumerate}[label=(\alph*)]
    \item a transition model describing the evolution of the internal, or latent, state $\bz$; and
    \item an observation model relating $\bz$ to the measured data $\bx$.
\end{enumerate}
Under the usual Markov and conditional-observation assumptions, these components are expressed as conditional probability distributions.
A predict--update recursion propagates the state distribution through the transition model and incorporates each new measurement through Bayes' rule.
Process and measurement noise are represented probabilistically, while discrepancies between predicted and observed responses inform the posterior update.
The resulting posterior distributions quantify uncertainty in the inferred quantities, subject to the adopted models and approximations.

Classical formulations include the \gls{KF} for linear systems, extended and unscented Kalman filters, particle filters, and related minimum-variance unbiased estimators.
Their formulations and applications to sequential estimation in structural dynamics are reviewed comprehensively in~\cite{tatsis2022sequential}.
Applications encompass system identification~\cite{khodarahmi2023review, rosafalco2024system, soal2019system, rosafalco2024ekf}, input estimation~\cite{impraimakis2022unscented, rodriguez2021state, song2023joint, liu2025model}, state estimation~\cite{impraimakis2025deep, vettori2023adaptive}, and parameter estimation~\cite{bi2023parameter, diaz2023new, li2022constrained, stoura2025identification}.
Coupling these estimators with appropriate \glspl{ROM} can enable operation in or near real time~\cite{agathos2022parametric, azzi2024enhanced, didyk2025towards, eftekhar2017online, porrello2024bayesian, tatsis2022hierarchical, vlachas2026utility}.

A central limitation in identification settings is that classical filters generally require prescribed transition and observation model structures.
In physics-based formulations, these follow from assumed governing equations, although states and selected physical parameters may remain uncertain~\cite{behmanesh2015hierarchical, jafari2025bayesian}.
In civil and mechanical engineering, \gls{FE} models commonly provide the dynamics~\cite{hollins2026projection, jang2017bayesian}, with parameter inference posed as Bayesian model updating or calibration~\cite{rocchetta2018line, lye2021sampling}.
The adequacy of the prescribed model therefore remains critical when governing mechanisms are partially understood or model-form uncertainty is substantial.

Advances in \gls{DL} offer a means of relaxing this requirement~\cite{pillonetto2025deep}.
Deep Bayesian formulations embed learnable representations within probabilistic state-space models and train model or inference components end to end~\cite{haarnoja2016backprop, jonschkowski2018differentiable, revach2022kalmannet}.
Examples include deep Kalman models~\cite{dahan2024uncertainty, dahan2025bayesian, krishnan2015deep} and other deep Bayesian filtering formulations~\cite{karl2016deep, wang2020survey, xiang2019online}.
Learning may affect the transition model, observation model, latent representation, or inference mechanism.
Variational approaches introduce a learned approximation to the posterior, whereas differentiable filtering approaches retain an algorithmic recursion, such as a Kalman predict--update scheme, within the learning procedure.
In both cases, inference remains coupled to the probabilistic dynamical model, distinguishing these formulations from neural representations without an explicit state-space structure.

The \gls{DMM} provides a useful starting point for understanding this construction~\cite{krishnan2017structured}.
Although its transition and observation mappings may be learned without prescribed physics, it already imposes probabilistic and temporal structure through its latent-state formulation.
The tutorial then examines how additional biases can be introduced: low-dimensional representations constrain the latent space, physical knowledge informs its coordinates and dynamics, and observation structure relates latent quantities to measured responses.
Inference introduces further assumptions through either a variational posterior or a filtering approximation.
These choices structure the learning problem in different ways and need not be introduced together.

Physics-enhanced examples include neural differential-equation formulations~\cite{lai2021structural}, the \gls{PgDMM}~\cite{liu2022physics}, and the Neural \gls{EKF}~\cite{liu2024neural}.
More broadly, learned estimation architectures have been investigated in Kalman filtering~\cite{haarnoja2016backprop}, variational state-space inference~\cite{krishnan2017structured}, particle filtering~\cite{karkus2018particle}, and unscented Kalman filtering~\cite{wigren2022nonlinear, wu2024physics}.
Reported comparisons with standalone recurrent, long short-term memory, and convolutional neural networks illustrate potential advantages of coupling learned representations with dynamical inference.
Such advantages, including robustness to model discrepancy, depend on the formulation, incorporated knowledge, and estimation setting.

Within this broader framework, the \gls{PgDMM}~\cite{liu2022physics} and Neural \gls{EKF}~\cite{liu2024neural} are examined in depth as complementary examples based on amortized variational inference and differentiable Kalman filtering, respectively.
The standard \gls{DMM} provides the reference against which the effects of additional physical structure can be assessed.
This progression is conceptual: the examples illustrate different combinations of modeling and inference biases, rather than a single sequence of architectural extensions.

For the physics-enhanced formulations considered here, a central ingredient is a low-order latent representation informed by \glspl{ROM}, which approximate the salient input--output behavior of high-fidelity models~\cite{benner2017model, padula2024brief, dhopeshwar2026two}.
This structure can give the latent state $\bz$ a physical interpretation and inform its transition dynamics, while learned components represent residual nonlinearities and model discrepancy.
The tutorial examines how these choices influence interpretability, uncertainty estimates, and predictive behavior, without assuming that physical structure alone guarantees accurate forecasts.

The formulations are demonstrated on a three-dimensional shear-frame benchmark with hysteretic Bouc--Wen joints under seismic excitation~\cite{vlachas2021two}.
The two identification settings cover response reconstruction and forecasting, latent-state recovery, and joint state and physical-parameter estimation within a common probabilistic perspective.
The comparative analysis also yields new observations on the distinction between measurement-conditioned reconstruction and open-loop forecasting, the relationship between latent-state interpretability and predictive fidelity, and differences in response and parameter-estimation robustness.
The aim is a coherent tutorial connecting first principles, Bayesian inference, and \gls{DL}, with practical guidance on formulation, training, evaluation, and deployment.

\paragraph{Positioning with respect to existing surveys and tutorials.}
The earlier tutorial on nonlinear Bayesian filtering for structural dynamics~\cite{tatsis2022sequential} covers classical Kalman-type and particle filtering methods for sequential state, input, and parameter estimation.
The survey on physics-enhanced machine learning for structural mechanics~\cite{Haywood2024} places Bayesian filtering, learned latent-state models, and physics-guided learning within the broader landscape of physics--data integration, but does not develop deep Bayesian filtering in detail.
General surveys on machine learning for system identification and hybrid physics-informed modeling likewise span multiple paradigms~\cite{ljung2020deep, magris2023bayesian, pillonetto2025deep, wang2020survey}, without providing the unified, step-by-step treatment pursued here.
Building on these foundations, this tutorial explains how probabilistic, structural, physical, and inference biases can be combined within learnable state-space formulations.
The \gls{DMM}, \gls{PgDMM}, and Neural \gls{EKF} make these principles concrete, supporting conceptual understanding and practical implementation without attempting an exhaustive survey of emerging methods.

The paper is organized as follows.
Section~\ref{sec:ProblemStatement} formulates dynamic state estimation and joint state and parameter estimation under sparse, noisy observations.
Section~\ref{sec:Methodology} develops the common probabilistic foundations, the \gls{DMM} baseline, the Neural \gls{EKF}, physically meaningful latent representations based on \glspl{ROM}, and the \gls{PgDMM}.
Section~\ref{sec:Applications} presents the shear-frame case study and evaluates response reconstruction, forecasting, latent-state recovery, and parameter estimation.
Section~\ref{sec:Discussion} distills the main findings, discusses training and deployment, and outlines future research directions.
\section{Problem Statement}\label{sec:ProblemStatement}

\subsection{System identification as an inverse problem}\label{sec:Problem_inverse}

A dynamical system can be represented abstractly through a state evolution model and an observation process,
\begin{align}
    \dot{\bz}(t) &= M\!\left(\bz(t),\bu(t);\bp\right), \label{eq:problem}\\
    \bx(t) &= H\!\left(\bz(t),\bu(t);\bp\right) + \be(t),
    \label{eq:problem_obs}
\end{align}
where $\bz$ denotes the system state, or more generally a latent representation of its dynamics, $\bu$ the external input, $\bx$ the available observations, and $\bp$ a vector of parameters governing the system response.
The functions $M$ and $H$ respectively describe the governing dynamics and the mapping from the underlying state to the measured quantities.
The dependence on $\bp$ is shown explicitly here and omitted where unnecessary for compactness; parameter inference is revisited in~\autoref{sec:Problem_parameter}.

Given discrete measurements, system identification may involve inference of the governing dynamics $M$, the observation mapping $H$, latent or unmeasured states, physical parameters, unknown inputs, or combinations thereof, depending on which components are known \emph{a priori}.
For the model-identification component of this broader inverse problem, one may consider observed input--output pairs $\left\{\bu_k,\bx_k\right\}_{k=1}^{N_k}$, with $\bx_k \in \mathbb{R}^{n_x}$ and $\bu_k \in \mathbb{R}^{n_u}$ denoting respectively the observed response and input excitation at discrete step $k$.
Given
\begin{enumerate}[label=(\alph*)]
    \item an abstract model space $\mathcal{M}$ containing admissible dynamical and observation models based on physical constraints and pre-existing knowledge, and
    \item the measurements obtained from the actual system,
\end{enumerate}
a deterministic system identification problem may be expressed as
\begin{align}\label{eq:inverse_problem}
    \left(\hat{M}^{\star},\hat{H}^{\star},\btheta^{\star}\right)
    =
    \argmin_{\hat{M},\,\hat{H}\in\mathcal{M},\,\btheta\in\Theta}
    \mathcal{F}\!\left(\hat{M},\hat{H},\btheta;\bx,\bu\right),
\end{align}
where $\hat{M}$ and $\hat{H}$ denote candidate transition and observation representations, $\btheta$ collects their unknown parameters, $\Theta$ is the corresponding parameter space, and $\mathcal{F}$ quantifies disagreement between the candidate model and the available observations.

Equation~\ref{eq:inverse_problem} provides a deterministic view of the identification problem.
In a Bayesian setting, the objective is instead to characterize posterior distributions over the unknown quantities, which may include latent states, inputs, physical parameters, and learnable components of the governing and observation models.
Deep Bayesian formulations differ in which of these components are prescribed or learned and in how the corresponding posterior distributions are represented and inferred.
For the representative formulations developed in~\autoref{sec:Methodology}, learning is carried out through variational objectives based on the \gls{ELBO}.

\subsection{State-space formulation with learnable models}\label{sec:Problem_statespace}

Adopting the observer formulation with learnable models and writing the problem in its discretized form in state-space leads to the following representation:
\begin{gather}
    \bz_k = f_{\nu}\left(\bz_{k-1}, \bu_k\right) + \bw_k 
    \quad \text{(transition)} \label{eq:transition} \\
    \bx_k = h_{\phi}\left(\bz_{k},\bu_k\right) + \be_k 
    \quad \text{(observation)} \label{eq:observation}
\end{gather}

where $k$ denotes the discretization step, so that the observation window comprises the $N_k$ discrete steps introduced above, $f_\nu$ and $h_\phi$ are nonlinear functions parameterized by $\nu$ and $\phi$ that govern the transition and observation models, $\bw_k$ and $\be_k$ are noise terms that account for process and measurement uncertainty, and $\bz \in \mathbb{R}^{n_z}$ is the latent state of the system that generates the observations $\bx$ through the observation model $h_\phi$.
This formulation underpins the deep Bayesian frameworks reviewed in this work, in which $f_\nu$ and $h_\phi$ are rendered learnable.
The parameters $\nu$ and $\phi$ may correspond to physical traits in the case of physics-based components, or to \gls{NN} weights and biases in the case of \gls{DL} variants.
Thus, inferring the unknown components of $M$ and $H$ in~\autoref{eq:inverse_problem} relates to learning the transition and observation models in Equations~\ref{eq:transition} and~\ref{eq:observation}, with $\nu$ and $\phi$ forming part of the overall set of learnable model parameters.

A key element of the physics-enhanced structural formulations considered in this tutorial is the choice of a low-dimensional latent state.
In this tutorial, this dimensional reduction is grounded in the theory of \glspl{ROM}, which provides a compact subspace that captures the dominant dynamics of the system.
The explicit construction and role of \glspl{ROM} as a physics-informed backbone for $\bz$ are detailed in~\autoref{sec:ROMback}.

\subsubsection*{Working assumptions.}
The treatment that follows relies on a small set of standard assumptions that we make explicit here:
\begin{enumerate}[label=(\alph*)]
    \item the noise terms $\bw_k$ and $\be_k$ are Gaussian, with $\bw_k \sim \mathcal{N}(0, \bm{Q}_k)$ and $\be_k \sim \mathcal{N}(0, \bm{R}_k)$
    \item the covariance matrices $\bm{Q}_k$ and $\bm{R}_k$ are learned from data, either as parameters or, in the generic \gls{DMM} and \gls{PgDMM}, as functions of the latent state
    \item the latent state $\bz$ follows a Markov process, so that, conditioned on the input, $\bz_k$ depends on the past only through $\bz_{k-1}$
    \item the observation sequence $\{\bx_k\}_{k=1}^{N_k}$ is obtained from a sparse monitoring scheme 
    \item the latent space dimension is low compared to the full-order model space of the system
\end{enumerate}

\subsection{Evaluation protocol}\label{sec:Problem_evaluation}
A learnable state-space model of the form in Equations~\ref{eq:transition} and~\ref{eq:observation} can be interrogated in several distinct ways, and these are routinely conflated in the literature.
We therefore distinguish here three complementary evaluations that are reported separately for the dynamic state estimation setting of this tutorial:
\begin{enumerate}[label=(\roman*)]
    \item \emph{posterior reconstruction}, in which the whole measurement window is assimilated and the response is reproduced from the resulting posterior over the latent state;
    \item \emph{open-loop forecasting}, in which a short measurement-conditioned prefix initializes the latent state and the remaining response is generated using only the learned dynamics and the known input; and
    \item \emph{structural state recovery}, in which the inferred latent coordinates are mapped back to physical quantities, namely displacements and velocities for the structural specializations used here, and compared with the hidden states of the data-generating system.
\end{enumerate}
Evaluations (i) and (ii) can differ by a factor of several on the same model, since a posterior that assimilates every measurement can mask a transition model that is locally inaccurate and only a rollout without measurement correction exposes that inaccuracy.
Evaluation (iii) in turn is what distinguishes a physically interpretable latent space from an arbitrary feature vector, and it is particularly informative for quantities that the model does not directly observe.
When physical parameters are inferred as well, as in~\autoref{sec:Problem_parameter}, a fourth evaluation is added, namely \emph{parameter recovery}, which compares the inferred parameter vector against the true one on held-out data.
Throughout, the axis of generalization is stated alongside the metric: the dynamic state estimation setting of~\autoref{sec:Problem_response} evaluates temporal generalization for a fixed physical system, whereas the joint dynamic state and parameter estimation setting of~\autoref{sec:Problem_parameter} evaluates generalization across parameter values.

\subsection{Dynamic state estimation}\label{sec:Problem_response}

We first consider the dynamic state estimation setting, in which the physical parameters of the system are fixed over the dataset under consideration and are neither appended to the observations nor represented by a dedicated latent block.
Given the measured input--response sequence $\left\{\bu_k,\bx_k\right\}_{k=1}^{N_k}$, the objective is to identify a probabilistic state-space model that supports the three evaluations (i)--(iii) of~\autoref{sec:Problem_evaluation}, with the structural state recovery of (iii) instantiated on displacement- and velocity-like coordinates.

This protocol evaluates temporal generalization for a fixed physical system and should not be interpreted as generalization to unseen parameter regimes.

\subsection{Joint dynamic state and parameter estimation}\label{sec:Problem_parameter}
Beyond response prediction, we additionally consider here the task of \emph{parameter estimation}, which seeks to infer a vector of physical parameters $\bp \in \mathbb{R}^{n_p}$ that govern the system response.
Such parameters typically correspond to system characteristics such as material properties or damping ratios, but may also characterize uncertain excitation or operating conditions.
These are assumed to remain time-invariant over each observation window while varying across operating regimes, system configurations, or realizations.
The general inference problem accordingly becomes one of recovering the latent dynamic state $\bz_k$ together with the parameter vector $\bp$ from the observed discrete sequence $\left\{\bu_k, \bx_k\right\}_{k=1}^{N_k}$, under the sparse-sensing assumption stated above.

A practical mechanism for handling parameter estimation, within the deep Bayesian frameworks reviewed in this work, is to augment the latent state with a dedicated parameter block,
\begin{align}\label{eq:latent_decomposition}
    \bz = \left[\bz^{\text{dyn}},\, \bz^{\text{par}}\right],
\end{align}
where $\bz^{\text{dyn}}$ captures the time-varying dynamics and $\bz^{\text{par}}$ is a latent-space representation of the physical parameter vector $\bp$, treated as near-constant over each observation window.

The mapping between $\bz^{\text{par}}$ and $\bp$ is learned jointly with the rest of the framework, and its precise treatment within the transition and inference mechanisms depends on the adopted framework, as detailed in~\autoref{sec:Methodology}.
\section{Methodology}\label{sec:Methodology}

Deep Bayesian filtering encompasses a family of probabilistic state-space formulations in which the transition and observation models are rendered learnable from data, while posterior inference may itself be learned or retained in algorithmic form.
Their common ingredients are therefore a probabilistic transition model governing the evolution of the latent state, an observation model relating this state to the available measurements, and a mechanism for posterior inference over the latent variables.
Different formulations principally differ in how these components are parameterized, how prior physical or structural knowledge is incorporated, and how posterior inference is carried out.

The present tutorial develops these common principles and then uses three representative formulations to make the discussion concrete:
\begin{itemize}
    \item The \gls{DMM} of~\cite{krishnan2017structured} provides a fully data-driven deep probabilistic state-space formulation and serves as a reference for amortized variational inference.
    \item The \gls{PgDMM}~\cite{liu2022physics} illustrates how physical knowledge can be incorporated within such a variational deep state-space model.
    \item The Neural \gls{EKF}~\cite{liu2024neural} represents a complementary class in which the transition and observation models remain learnable but latent-state inference is performed through differentiable Kalman filtering recursions.
\end{itemize}
The latter two formulations are developed in greater depth as representative structured, physics-aware realizations, rather than as an exhaustive enumeration of deep Bayesian filtering methodologies.
Template software implementations are available for the \gls{PgDMM}\footnote[1]{\url{https://github.com/liouvill/PgDMM}} and the Neural \gls{EKF}\footnote[2]{\url{https://github.com/liouvill/NeuralEKF}}.

The methodological discussion is organized around the following common building blocks.
We first introduce the shared temporal probabilistic state-space structure and the variational learning principle underlying the considered formulations.
The \gls{DMM} is then presented as a canonical data-driven realization of this structure, providing the black-box reference against which the role of additional physical structure can be understood.
We subsequently turn to mechanisms for introducing such structure, first through the Neural \gls{EKF}, which retains learnable models but performs inference with the recursions of a classical filter, and then through \glspl{ROM} as one means of embedding low-order physical knowledge into the latent dynamics, on which the \gls{PgDMM} builds.
The Neural \gls{EKF} and the \gls{PgDMM} are thus developed as two complementary physics-aware formulations, differing in how physical knowledge is incorporated into the state-space representation and how posterior inference is performed. For each framework, we first introduce the general transition, observation, and inference mechanisms, before presenting their structural specializations for \emph{dynamic state} estimation and \emph{joint dynamic state and parameter} estimation. These correspond to the problems formulated in~\autoref{sec:Problem_response} and~\autoref{sec:Problem_parameter}, with numerical demonstrations provided in~\autoref{sec:responseOnly} and~\autoref{sec:responseParam}, respectively.

\subsection{Temporal variational state-space skeleton}\label{sec:tVAE}

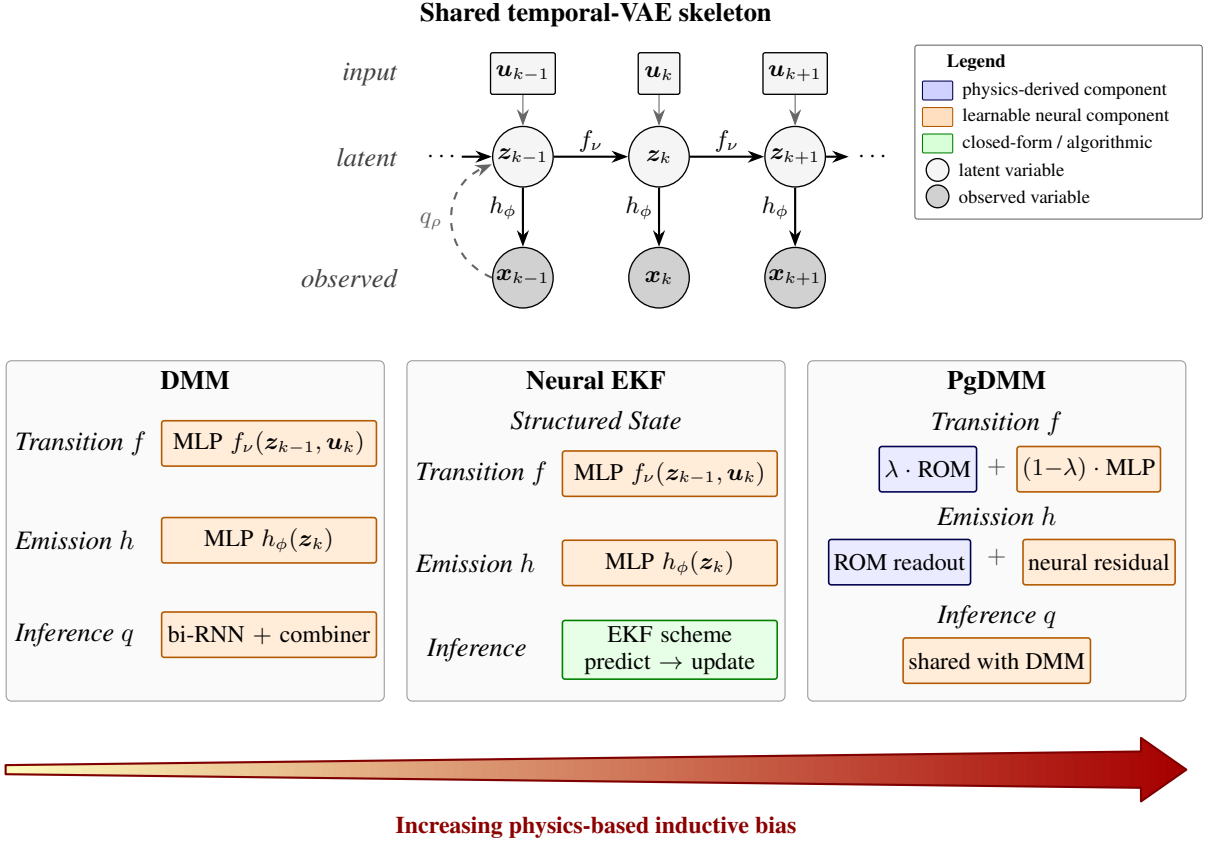
\begin{figure}[!ht]
\centering
\begin{tikzpicture}[
    >={Stealth[length=2.2mm,width=1.6mm]},
    font=\footnotesize,
    every node/.style={font=\footnotesize},
    zvar/.style={
        circle, draw=black, line width=0.5pt,
        fill=black!4,
        minimum size=8mm, inner sep=0pt,
    },
    xvar/.style={
        circle, draw=black, line width=0.5pt,
        fill=black!18,
        minimum size=8mm, inner sep=0pt,
    },
    uvar/.style={
        rectangle, draw=black, line width=0.5pt,
        rounded corners=0.8pt,
        fill=black!4,
        minimum size=5.5mm, inner sep=2pt,
    },
    phyblock/.style={
        rectangle, draw=black!70!blue, line width=0.6pt,
        rounded corners=1pt,
        fill=blue!8,
        align=center,
        minimum height=6mm, inner sep=2pt,
    },
    nnblock/.style={
        rectangle, draw=orange!70!black, line width=0.6pt,
        rounded corners=1pt,
        fill=orange!15,
        align=center,
        minimum height=6mm, inner sep=2pt,
    },
    infblock/.style={
        rectangle, draw=green!50!black, line width=0.6pt,
        rounded corners=1pt,
        fill=green!10,
        align=center,
        minimum height=6mm, inner sep=2pt,
    },
    panelbg/.style={
        rectangle, draw=black!40, line width=0.4pt,
        rounded corners=2pt,
        fill=black!2,
    },
    paneltitle/.style={font=\bfseries},
    rowlab/.style={font=\itshape, text=black!75, align=right},
    caption/.style={font=\scriptsize\itshape, text=black!70, align=center},
    legendbox/.style={
        rectangle, draw=black, line width=0.4pt,
        rounded corners=0.5pt,
        minimum width=4mm, minimum height=2.5mm,
        inner sep=0pt,
    },
]

\begin{scope}[local bounding box=skeleton]

    \def\colsep{18mm}

    \node[zvar] (zm) at (0,0) {$\bm{z}_{k-1}$};
    \node[zvar] (zt) at (\colsep,0) {$\bm{z}_{k}$};
    \node[zvar] (zp) at (2*\colsep,0) {$\bm{z}_{k+1}$};

    \node[left=3mm of zm]  (zdotsL) {$\cdots$};
    \node[right=3mm of zp] (zdotsR) {$\cdots$};

    \node[xvar] (ym) at (0,-16mm)         {$\bm{x}_{k-1}$};
    \node[xvar] (yt) at (\colsep,-16mm)   {$\bm{x}_{k}$};
    \node[xvar] (yp) at (2*\colsep,-16mm) {$\bm{x}_{k+1}$};

    \node[uvar] (um) at (0,11mm)         {$\bm{u}_{k-1}$};
    \node[uvar] (ut) at (\colsep,11mm)   {$\bm{u}_{k}$};
    \node[uvar] (up) at (2*\colsep,11mm) {$\bm{u}_{k+1}$};

    \draw[->, thick] ([xshift=-1mm]zdotsL.east) -- (zm.west);
    \draw[->, thick] (zm.east) -- node[above=-0.7mm]{$f_\nu$} (zt.west);
    \draw[->, thick] (zt.east) -- node[above=-0.7mm]{$f_\nu$} (zp.west);
    \draw[->, thick] (zp.east) -- (zdotsR.west);

    \draw[->, thick] (zm.south) -- node[yshift=1.25mm,left=-0.5mm]{$h_\phi$} (ym.north);
    \draw[->, thick] (zt.south) -- node[yshift=1.25mm,left=-0.5mm]{$h_\phi$} (yt.north);
    \draw[->, thick] (zp.south) -- node[yshift=1.25mm, left=-0.5mm]{$h_\phi$} (yp.north);

    \draw[->, thin, black!60] (um.south) -- (zm.north);
    \draw[->, thin, black!60] (ut.south) -- (zt.north);
    \draw[->, thin, black!60] (up.south) -- (zp.north);

    \draw[->, dashed, thick, black!60]
        (ym.west) .. controls +(-7mm,3mm) and +(-7mm,-3mm) .. ([yshift=-1mm]zm.west) node[midway, left] {$q_\rho$};

    \node[rowlab, anchor=east] at ($(zdotsL.west)+(-1.5mm,11mm)$)  {input};
    \node[rowlab, anchor=east] at ($(zdotsL.west)+(-1.5mm,0mm)$)  {latent};
    \node[rowlab, anchor=east] at ($(zdotsL.west)+(-1.5mm,-16mm)$){observed};

\end{scope}

\node[anchor=north, font=\scriptsize, text=black!80]
    at ($(skeleton.south)+(0,-1.5mm)$) (zsplit)
    {};

\node[paneltitle]
    at ($(skeleton.north)+(0mm,5mm)$)
    {Shared temporal-VAE skeleton};

\begin{scope}[shift={($(skeleton.north east)+(5mm,-5mm)$)}]

    \draw[draw=black!60, thin, rounded corners=1pt] (-3mm, 6mm) rectangle (35mm, -17mm);
    \node[anchor=west, font=\scriptsize\bfseries] (ltitle) at (0mm, 3.75mm) {Legend};

    \node[legendbox, fill=blue!18, draw=black!70!blue] (l1) at (0,0) {};
    \node[anchor=west, font=\scriptsize] at (l1.east) {physics-derived component};
    \node[legendbox, fill=orange!25, draw=orange!70!black] (l2) at (0,-3.5mm) {};
    \node[anchor=west, font=\scriptsize] at (l2.east) {learnable neural component};
    \node[legendbox, fill=green!15, draw=green!50!black] (l3) at (0,-7mm) {};
    \node[anchor=west, font=\scriptsize] at (l3.east) {closed-form / algorithmic};
    \node[circle, draw=black, line width=0.5pt, fill=black!4,
          minimum size=3mm, inner sep=0pt] (l4) at (0,-10.5mm) {};
    \node[anchor=west, font=\scriptsize] at (l4.east) {latent variable};
    \node[circle, draw=black, line width=0.5pt, fill=black!18,
          minimum size=3mm, inner sep=0pt] (l5) at (0,-14mm) {};
    \node[anchor=west, font=\scriptsize] at (l5.east) {observed variable};
\end{scope}

\coordinate (rowAnchor) at ($(zsplit.south)+(0,-3mm)$);

\def\pW{50mm}
\def\pH{45mm}
\def\pSep{53mm}   %

\begin{scope}[shift={($(rowAnchor)+(-\pSep,-\pH/2)$)}, local bounding box=dmmpanel]
    \node[panelbg, minimum width=\pW, minimum height=\pH] (dmmbg) at (0,0) {};
    \node[paneltitle, anchor=north, inner sep=1pt]
        at ([yshift=-1mm]dmmbg.north)
        {DMM};

    \node[anchor=west, font=\itshape] at ($(dmmbg.north west)+(0mm,-11mm)$)
        (dmmtr_lab) {Transition $f$};
    \node[nnblock, anchor=east, text width=27mm]
        at ($(dmmbg.north east)+(-1mm,-11mm)$) (dmmtr)
        {MLP $f_\nu(\bm{z}_{k-1},\bm{u}_k)$};

    \node[anchor=west, font=\itshape] at ($(dmmtr_lab.south west)+(0,-10mm)$)
        (dmmem_lab) {Emission $h$};
    \node[nnblock, anchor=east, text width=27mm]
        at ($(dmmem_lab.east -| dmmtr.east)$) (dmmem)
        {MLP $h_\phi(\bm{z}_k)$};

    \node[anchor=west, font=\itshape] at ($(dmmem_lab.south west)+(0,-10mm)$)
        (dmminf_lab) {Inference $q$};
    \node[nnblock, anchor=east, text width=27mm]
        at ($(dmminf_lab.east -| dmmtr.east)$) (dmminf)
        {bi-RNN $+$ combiner};

\end{scope}

\begin{scope}[shift={($(rowAnchor)+(0,-\pH/2)$)}, local bounding box=ekfpanel]
\node[panelbg, minimum width=\pW, minimum height=\pH] (ekfbg) at (0,0) {};
    \node[paneltitle, anchor=north, inner sep=1pt]
        at ([yshift=-1mm]ekfbg.north)
        {Neural EKF};

    \node[font=\itshape, anchor=north]
        at ($(ekfbg.north)+(0,-5.5mm)$) (pgtr_lab) {Structured State};

    \node[anchor=west, font=\itshape] at ($(ekfbg.north west)+(0mm,-15mm)$)
        (ektr_lab) {Transition $f$};
    \node[nnblock, anchor=east, text width=27mm]
        at ($(ekfbg.north east)+(-1mm,-15mm)$) (ektr)
        {MLP $f_\nu(\bm{z}_{k-1},\bm{u}_k)$};

    \node[anchor=west, font=\itshape] at ($(ektr_lab.south west)+(0,-9mm)$)
        (ekem_lab) {Emission $h$};
    \node[nnblock, anchor=east, text width=27mm]
        at ($(ekem_lab.east -| ektr.east)$) (ekem)
        {MLP $h_\phi(\bm{z}_k)$};

    \node[anchor=west, font=\itshape] at ($(ekem_lab.south west)+(1.5mm,-9mm)$)
        (ekinf_lab) {Inference};
    \node[infblock, anchor=east, text width=27mm]
        at ($(ekinf_lab.east -| ektr.east)$) (ekinf)
        {EKF scheme \\ predict $\to$ update};
 \end{scope}

\begin{scope}[shift={($(rowAnchor)+(\pSep,-\pH/2)$)}, local bounding box=pgpanel]
    \node[panelbg, minimum width=\pW, minimum height=\pH] (pgbg) at (0,0) {};
    \node[paneltitle, anchor=north, inner sep=1pt]
        at ([yshift=-1mm]pgbg.north)
        {PgDMM};

    \node[font=\itshape, anchor=north]
        at ($(pgbg.north)+(0,-5.5mm)$) (pgtr_lab) {Transition $f$};

    \node[font=\footnotesize, text=black!70]
        at (pgtr_lab.south) [yshift=-3.0mm]{$+$};

    \node[phyblock, anchor=north east, align=center]
        at ($(pgtr_lab.east)+(-12.5mm,-3mm)$) (pgtrP)
        {$\lambda$ $\cdot$ ROM};
    \node[nnblock, anchor=north west, align=center]
        at ($(pgtr_lab.west)+(+12.5mm,-3mm)$) (pgtrN)
        {$(1{-}\lambda)$ $\cdot$ MLP};

    \node[font=\itshape, anchor=north]
        at ($(pgtr_lab.south)+(0,-7mm)$) (pgem_lab) {Emission $h$};

    \node[font=\footnotesize, text=black!70]
        at (pgem_lab.south) [yshift=-3.0mm]{$+$};

    \node[phyblock, anchor=north east, align=center]
        at ($(pgem_lab.east)+(-12.5mm,-3mm)$) (pgemP)
        {ROM readout};
    \node[nnblock, anchor=north west, align=center]
        at ($(pgem_lab.west)+(+12.5mm,-3mm)$) (pgemN)
        {neural residual};

    \node[font=\itshape, anchor=north]
        at ($(pgtr_lab.south)+(0,-20mm)$) (pginf_lab) {Inference $q$};
    \node[nnblock, align=center]
        at ($(pginf_lab.south)+(0,-3mm)$) (pginf)
        {shared with DMM};
\end{scope}

\coordinate (biasL) at ([yshift=-9mm]dmmbg.south west);
\coordinate (biasR) at ([yshift=-9mm]pgbg.south east);

\path[
    shade,
    shading=axis,
    left color=yellow!25,
    right color=red!65!black,
    draw=red!50!black,
    line width=0.8pt,
    line join=round
]
    ($(biasL)+(0,-0.6mm)$)
    --
    ($(biasR)+(-6mm,-2.4mm)$)
    --
    ($(biasR)+(-6mm,-4.2mm)$)
    --
    (biasR)
    --
    ($(biasR)+(-6mm,4.2mm)$)
    --
    ($(biasR)+(-6mm,2.4mm)$)
    --
    ($(biasL)+(0,0.6mm)$)
    -- cycle;

\node[
    below=5mm of $(biasL)!0.5!(biasR)$,
    font=\footnotesize\bfseries,
    text=red!55!black
]
    {Increasing physics-based inductive bias};

\end{tikzpicture}
\caption{Common probabilistic state-space skeleton and method-specific realizations of the three deep Bayesian filtering frameworks considered in this work. The top panel shows the latent state $\bz_k$ evolving under the transition model $f_\nu$ and generating observations $\bx_k$ through the emission model $h_\phi$, with the dynamics driven by inputs $\bu_k$. Inference recovers the latent state from the observation sequence through an approximate posterior $q$. The bottom panels arrange the three frameworks according to increasing physics-based inductive bias. The \gls{DMM} constitutes the fully data-driven baseline, with learned transition, emission, and amortized inference models and no prescribed physical organization of the latent state. The Neural \gls{EKF} introduces an intermediate level of structure by replacing the amortized encoder with the \gls{EKF} predict-update recursion and, in its structural specialization, by organizing the latent state into displacement- and velocity-like coordinates, while retaining learnable transition and emission models. The \gls{PgDMM} introduces stronger physics-based bias by embedding a parameter-dependent \gls{ROM} within the transition and emission models, while retaining the amortized inference architecture of the \gls{DMM}.}
\label{fig:tVAE}
\end{figure}

To place the representative formulations on a common footing, we first introduce the temporal probabilistic structure that underlies them, as illustrated in~\autoref{fig:tVAE}.
At each discrete step, the latent state $\bz_k$ evolves according to a transition density
$p_\nu(\bz_k|\bz_{k-1},\bu_k)$ and generates the corresponding observation through an observation, or emission, density
$p_\phi(\bx_k|\bz_k,\bu_k)$.
These densities are the probabilistic counterparts of the transition and observation functions $f_\nu$ and $h_\phi$ introduced in Equations~\ref{eq:transition} and~\ref{eq:observation}.
Under the Gaussian setting adopted here,
\begin{align}
p_\nu(\bz_k|\bz_{k-1},\bu_k)
&=
\mathcal{N}\!\left(
f_\nu(\bz_{k-1},\bu_k),\,\bm{Q}_k
\right),\\
p_\phi(\bx_k|\bz_k,\bu_k)
&=
\mathcal{N}\!\left(
h_\phi(\bz_k,\bu_k),\,\bm{R}_k
\right),
\end{align}
where the covariance descriptions may be represented by global parameters or parameterized as functions of the latent state, depending on the adopted formulation.

Completing this temporal state-space description is the posterior distribution over the latent trajectory,
\begin{align}
q\!\left(
\bz_{1:N_k}\mid
\bx_{1:N_k},\bu_{1:N_k}
\right),
\end{align}
which recovers the latent state from the available observations.
The three representative formulations differ principally in how the transition, observation, and posterior-inference components of this common temporal skeleton are parameterized.
In the \gls{DMM} and \gls{PgDMM}, posterior inference is amortized through a learnable model
$q_\rho(\bz_{1:N_k}|\bx_{1:N_k},\bu_{1:N_k})$, yielding the familiar temporal \gls{VAE} construction.
In the Neural \gls{EKF}, the same inference role is instead fulfilled algorithmically through the differentiable \gls{EKF} predict-update recursions, without introducing a separate inference network.

Despite this distinction, all three formulations can be expressed within a common variational learning perspective.
Denoting the parameters of the generative model by $\btheta=[\nu;\phi]$, an \gls{ELBO} on the marginal log-likelihood can be written generically as
\begin{equation}\label{eq:elbogeneric}
\log p_{\btheta}\left(\bx_{1:N_k}\mid\bu_{1:N_k}\right)
\geq
\mathbb{E}_{q}\!\left[
\log p_\phi\left(
\bx_{1:N_k}\mid\bz_{1:N_k},\bu_{1:N_k}
\right)
\right]
-
\mathrm{KL}\!\left[
q\left(
\bz_{1:N_k}\mid\bx_{1:N_k},\bu_{1:N_k}
\right)
\,\|\, 
p_\nu\left(
\bz_{1:N_k}\mid\bu_{1:N_k}
\right)
\right]
\end{equation}
For the \gls{DMM} and \gls{PgDMM}, $q=q_\rho$ is parameterized by a dedicated inference network and its parameters $\rho$ are optimized jointly with the generative model.
For the Neural \gls{EKF}, $q$ is instead induced algorithmically by the filtering and smoothing recursions, without introducing a separate inference network or associated parameters.

The first term in~\autoref{eq:elbogeneric} promotes fidelity to the observations, while the \gls{KL} divergence regularizes the inferred posterior toward the latent dynamics represented by the transition model.
This common representation provides a convenient basis for comparing how different deep Bayesian formulations instantiate three central components: the transition density $p_\nu$, the observation density $p_\phi$, and the posterior inference mechanism $q$.
\autoref{fig:tVAE} summarizes these choices.

Specifically, the \gls{DMM} parameterizes the transition, observation, and inference components using \glspl{NN} and introduces no explicit physical bias, serving as the data-driven reference.
The \gls{PgDMM} introduces physics-derived structure into the transition and observation models while retaining the amortized inference construction of the \gls{DMM}.
The Neural \gls{EKF} likewise renders the transition and observation models learnable, but replaces the dedicated inference network with the standard \gls{EKF} predict-update recursions.

\subsection{The Deep Markov Model (DMM)}\label{sec:DMM}

We first instantiate the common temporal structure introduced above in its fully data-driven form through the \gls{DMM}.
In this formulation, the transition, observation, and amortized inference models are all parameterized from data, without imposing an explicit physical structure on the latent space.
Following the state-space formulation in Equations~\ref{eq:transition} and~\ref{eq:observation}, the joint distribution over latent states $\bz_{1:N_k}$ and observations $\bx_{1:N_k}$ is encoded by a marginal likelihood
\begin{equation}\label{eq:prob_DMM}
p_{\btheta}(\bx_{1:N_k}) = \int p_{\phi}(\bx_{1:N_k}|\bz_{1:N_k})\, p_{\nu}(\bz_{1:N_k})\, \mathrm{d}\bz_{1:N_k}
\end{equation}
where the subscript $\bullet_{1:N_k}$ denotes a sequence over the full observation window of $N_k$ discretization steps.
The transition function $f_{\nu}$ in~\autoref{eq:transition} is represented by its probabilistic counterpart $p_\nu$, parameterized by $\nu$, while the observation function $h_{\phi}$ in~\autoref{eq:observation} is represented by $p_\phi$, parameterized by $\phi$.
The full generative model is therefore parameterized by $\btheta = [\nu;\, \phi]$.
Conditional independence assumptions consistent with the Markov property are imposed on the transition and observation models, which factorize as
\begin{gather}
p_{\nu}(\bz_{1:N_k}) = p_{\nu}(\bz_0) \prod_{k=1}^{N_k} p_{\nu}(\bz_k|\bz_{k-1}) \quad \text{(transition)} \label{eq:transitionDMM} \\
p_{\phi}(\bx_{1:N_k}|\bz_{1:N_k}) = \prod_{k=1}^{N_k} p_{\phi}(\bx_k|\bz_k) \quad \text{(observation)} \label{eq:observationDMM}
\end{gather}
where the transition model is assumed Markovian, so that the current latent state depends only on the previous one, and the observation model assumes that the current observation can be recovered from the current latent state alone.
The generative model is time-invariant, so the parameters $\nu$ and $\phi$ do not depend on the step $k$, and the dependence on the input $\bu_k$ is omitted for compactness from the densities of this section and of~\autoref{sec:PgDMM}.

The transition and observation densities are cast as Gaussian distributions whose means and covariances are emitted by feed-forward \glspl{NN}:
\begin{gather}
p_\nu\left(\bz_k|\bz_{k-1}\right) = \mathcal{N}\!\left(\bmu_\nu(\bz_{k-1}),\, \bm{\Sigma}_\nu(\bz_{k-1})\right), \\
p_\phi\left(\bx_k|\bz_k\right) = \mathcal{N}\!\left(\bmu_\phi(\bz_{k}),\, \bm{\Sigma}_\phi(\bz_{k})\right),
\end{gather}
with $\left[\bmu_\nu, \bm{\Sigma}_\nu\right]$ and $\left[\bmu_\phi, \bm{\Sigma}_\phi\right]$ produced by independent \glspl{NN}.
The inference model $q_\rho\left(\bz_{1:N_k}|\bx_{1:N_k}, \bu_{1:N_k}\right)$ is parameterized by a bidirectional recurrent network that processes the observation sequence in both temporal directions, followed by a combiner network that fuses the resulting hidden states with the previous latent sample and the current control input.
The recurrent backbone employed here is a Gated Recurrent Unit (GRU), a lightweight variant of recurrent \glspl{NN} that gates the flow of past information into the hidden state and accordingly mitigates vanishing gradients without the parameter overhead of a long short-term memory cell.

Training proceeds by maximizing the \gls{ELBO} in~\autoref{eq:elbogeneric}, with the expectation evaluated through reparametrized Monte Carlo samples and the \gls{KL} term computed in closed form for the Gaussian factorization adopted here.

\begin{practicaltip}
The choice of variational family affects both the expressiveness and computational tractability of posterior inference. Under the Gaussian assumptions adopted here, the \gls{KL} divergence in the \gls{ELBO} of~\autoref{eq:elbogeneric} can be evaluated analytically. More expressive variational families, such as normalizing flows based on Inverse Autoregressive Flows, generally require Monte Carlo estimation of the \gls{KL} term using the variational log-density and the prior log-density, introducing additional computational cost and estimator variance. The Gaussian formulation is retained throughout this work for computational tractability.
\end{practicaltip}

The \gls{DMM} therefore inherits the flexibility of deep generative models, but its latent coordinates are not endowed with an intrinsic structural correspondence to the underlying physical system. A first step toward a more structured representation is to retain learnable transition and observation models while replacing amortized inference with the recursive Bayesian inference mechanism of a classical filter. The Neural \gls{EKF}, introduced next, follows this route and additionally allows the latent state to be organized according to the displacement--velocity structure of a second-order mechanical system.

\subsection{The Neural Extended Kalman Filter (Neural EKF)}\label{sec:NEKF}

The DMM introduced above represents the fully data-driven end of the deep Bayesian filtering spectrum: its transition, observation, and amortized inference models are learned from data, without assigning a prescribed physical meaning to the latent coordinates. The Neural \gls{EKF} introduces additional inductive structure in two respects. First, it replaces the amortized inference network with the predict-update recursion of the classical \gls{EKF}, while retaining learnable transition and observation models. Second, for the structural response specialization considered later in this section, its latent state is explicitly organized into displacement- and velocity-like coordinates whose evolution follows the kinematic structure of a second-order mechanical system. It therefore retains substantial data-driven flexibility while providing a more structured and physically interpretable state-space representation than the DMM.

The underlying idea was originally presented in~\cite{li2021replay}, where a replay-overshooting scheme was proposed, and was further refined in~\cite{liu2024neural} in the Neural \gls{EKF} formulation adopted here. Unlike the DMM, which uses a dedicated amortized inference network, the Neural \gls{EKF} obtains the latent-state posterior through the differentiable \gls{EKF} predict-update recursion, followed, where appropriate, by Rauch-Tung-Striebel backward smoothing. End-to-end training can therefore focus on the learnable transition and observation models without introducing a separate inference network. The transition and observation functions $f_\nu$ and $h_\phi$ in Equations~\ref{eq:transition} and~\ref{eq:observation} are accordingly rendered learnable, relaxing the requirement for explicitly prescribed observer models in conventional Bayesian filtering.

Throughout this section, the notation $\bullet_{i|j}$ denotes a quantity at time step $i$ conditioned on observations up to step $j$. Accordingly, $\bullet_{k|k-1}$ represents the predicted estimate prior to measurement assimilation, $\bullet_{k|k}$ the filtered estimate following the measurement update, and $\bullet_{k|N_k}$ the smoothed estimate incorporating information from the entire observation sequence.

\subsubsection{Background: EKF filtering and smoothing}\label{sec:KF}

To make explicit which part of the Neural \gls{EKF} is retained from classical Bayesian filtering, we first briefly recall the standard \gls{EKF} prediction and measurement-update recursion, followed by the backward smoothing step used here.  
The \gls{EKF} relies on the observers in Equations~\ref{eq:transition} and~\ref{eq:observation}, although the nonlinear functions $f$ and $h$ are assumed known and unparameterized.
The setup further assumes that the posterior of the latent state $\bz_k$ given the observations and inputs up to step $k$ is Gaussian, and is computed in closed form by the \gls{EKF}:
\begin{equation}
q(\bz_k|\bx_{1:k}, \bu_{1:k}) = \mathcal{N}\!\left(\bmu_{k|k},\, \bm{\Sigma}_{k|k}\right)
\end{equation}
where $\bmu_{k|k}$ and $\bm{\Sigma}_{k|k}$ are the mean and covariance of the inferred posterior. 
The \gls{EKF} conducts approximate inference by linearizing the nonlinear equations, with the posterior obtained via iterative updates of a predict-and-update process.
First, the prior is computed based on previous observations:
\begin{gather}
q(\bz_k|\bx_{1:k-1}, \bu_{1:k}) = \mathcal{N}\!\left(\bmu_{k|k-1},\, \bm{\Sigma}_{k|k-1}\right)\\
\bmu_{k|k-1} = f(\bmu_{k-1|k-1}, \bu_k) \label{eq:first}\\
\bm{\Sigma}_{k|k-1} = \bm{A}_{k-1}\, \bm{\Sigma}_{k-1|k-1}\, \bm{A}_{k-1}^T + \bm{Q}_{k-1}\\
\bm{A}_{k-1} = \frac{\partial f\left(\bmu_{k-1|k-1}, \bu_k\right)}{\partial \bmu_{k-1|k-1}}
\end{gather}
The update step then incorporates the current observation $\bx_k$:
\begin{gather}
\bm{G}_k = \bm{\Sigma}_{k|k-1}\, \bm{C}_k^T \left(\bm{C}_k\, \bm{\Sigma}_{k|k-1}\, \bm{C}_k^T + \bm{R}_k\right)^{-1}\\
\bmu_{k|k} = \bmu_{k|k-1} + \bm{G}_k\left[\bx_k - h(\bmu_{k|k-1})\right]\\
\bm{\Sigma}_{k|k} = \left(\bm{I} - \bm{G}_k\, \bm{C}_k\right) \bm{\Sigma}_{k|k-1}\\
\bm{C}_k = \frac{\partial h\left(\bmu_{k|k-1}\right)}{\partial \bmu_{k|k-1}}\label{eq:last}
\end{gather}
The Jacobian matrices $\bm{A}_k$ and $\bm{C}_k$ provide local linear approximations of the nonlinear functions for the purpose of propagating the covariances, and $\bm{G}_k$ denotes the Kalman gain, written here as $\bm{G}$ rather than the customary $\bm{K}$ to reserve the latter for structural stiffness.
 
The standard \gls{EKF} is typically deployed as a recursive filtering method, assuming a sequence of observations is available.
The posterior estimates can be further refined by sweeping back over the entire sequence, a process termed smoothing.
The Rauch-Tung-Striebel algorithm~\cite{rauch1965maximum} is a suitable scheme for this purpose: it operates backwards after the \gls{EKF} filtering pass is completed and updates the posteriors as
\begin{gather}
\bm{G}_k^s = \bm{\Sigma}_{k|k}\, \bm{A}_k^T \left(\bm{\Sigma}_{k+1|k}\right)^{-1}\\
\bmu_{k|N_k} = \bmu_{k|k} + \bm{G}_k^s\left(\bmu_{k+1|N_k} - \bmu_{k+1|k}\right)\\
\bm{\Sigma}_{k|N_k} = \bm{\Sigma}_{k|k} + \bm{G}_k^s\left(\bm{\Sigma}_{k+1|N_k} - \bm{\Sigma}_{k+1|k}\right)\left(\bm{G}_k^s\right)^T
\end{gather}
where the intermediate quantities are already available from the filtering pass in Equations~\ref{eq:first}--\ref{eq:last}.
 
\begin{shadednote}
    Numerical stability of the \gls{EKF} relies on preserving the symmetry and positive semi-definiteness of the propagated and updated covariance matrices, $\bm{\Sigma}_{k|k-1}$ and $\bm{\Sigma}_{k|k}$. In differentiable implementations, finite-precision arithmetic and gradient-based optimization may introduce numerical inconsistencies that violate these properties, potentially causing Cholesky factorization failures and undefined values (\texttt{NaN}) during training. Practical safeguards include covariance symmetrization, diagonal regularization $\bm{\Sigma} \leftarrow \bm{\Sigma} + \epsilon\bm{I}$ with a small $\epsilon>0$, and numerically stable covariance updates. If factorization failures persist, a controlled reinitialization of the affected covariance may prevent training from terminating, although repeated failures should prompt further investigation of the learned dynamics and covariance parameterization.
\end{shadednote}

These recursions define the structured inference mechanism retained by the Neural \gls{EKF}. The remaining ingredients are the transition and observation functions entering the prediction and update steps. Whereas these functions are prescribed in the classical \gls{EKF}, the Neural \gls{EKF} renders them learnable, as described next. The additional mechanical structuring of the latent state adopted for the structural examples is introduced separately in \autoref{sec:NEKFresponse}.

\subsubsection{Transition and observation models}

With the classical inference mechanism established above, the Neural \gls{EKF} is obtained by replacing the prescribed transition and observation functions with learnable counterparts.
The standard \gls{EKF} requires explicit knowledge of the nonlinear transition $f$ and observation function $h$; the Neural \gls{EKF} lifts this restriction by representing $f_\nu$ and $h_\phi$ with \glspl{NN}, trained end-to-end through the differentiable filtering and smoothing recursions.
The framework adopts the observer formulation in Equations~\ref{eq:transition} and~\ref{eq:observation}, where $f_\nu$ and $h_\phi$ are the learnable functions governing the transition and observation models, both parameterized by \glspl{NN} with parameters $\btheta = [\nu;\, \phi]$.
The process and observation noise sources $\bw$ and $\be$ are assumed Gaussian with time-invariant, learnable covariances.
 
For inference, the Neural \gls{EKF} follows the standard \gls{EKF} described in~\autoref{sec:KF}.
Once a Neural \gls{EKF} model is trained, the learned functions $f_\nu$ and $h_\phi$ are readily available and can be plugged into Equations~\ref{eq:first}--\ref{eq:last} to perform closed-form inference.
Unlike the \gls{DMM}, no separate inference network is parameterized: the variational posterior $q$ is computed in closed form by the \gls{EKF}, and consequently $q$ inherits the same parameters $\btheta$ as the generative model.
 
\subsubsection{Evidence lower bound objective}
 
With the \gls{EKF} conducting inference, the parameters to be learned are summarized in $\btheta$ and include the parameters of the \glspl{NN} used to model the transition $f_\nu$ and observation $h_\phi$ functions.
In addition, the initial values $\bmu_{0|0}$, $\bm{\Sigma}_{0|0}$ and the covariances $\bm{Q}, \bm{R}$ are also learned.
During training, the Neural \gls{EKF} integrates variational inference principles, with the \gls{EKF} algorithm performing the inference task.
Given an inference distribution $q_{\btheta}\left(\bz_k|\bx, \bu\right)$ obtained in closed form from the \gls{EKF} and assuming Markovian dynamics, the \gls{ELBO} can be expressed in factorized form as
\begin{align}
\mathcal{L}(\btheta;\, \bx) = \sum_{k=1}^{N_k} \Bigg( & \mathbb{E}_{q_{\btheta}(\bz_k | \bx,\bu)}\!\left[\log p_{\phi}(\bx_k|\bz_k)\right] \nonumber \\
& - \mathbb{E}_{q_{\btheta}(\bz_{k-1}|\bx,\bu)}\!\left[\mathrm{KL}\!\left(q_{\btheta}(\bz_k|\bx, \bu) \,\|\, p_{\nu}(\bz_k|\bz_{k-1}, \bu_k)\right)\right] \Bigg) \label{eq:elbo1}
\end{align}
where the $\bullet_{1:N_k}$ subscripts have been dropped from $\bx$ and $\bu$ for compactness.
Since the posterior $q_{\btheta}\left(\bz_k|\bx, \bu\right)$ is computed in closed form, the distributions in~\autoref{eq:elbo1} can be evaluated explicitly given $\bz_k \sim \mathcal{N}\!\left(\bmu_{k|N_k},\, \bm{\Sigma}_{k|N_k}\right)$.
The detailed derivation is provided in the original publication~\cite{liu2024neural}.
We note that the log-probability and \gls{KL}-divergence terms can be computed analytically because all the distributions involved are Gaussian.
After substituting the \gls{EKF} and smoothing recursions of~\autoref{sec:KF}, and evaluating the likelihood term with the predictive covariance $\bm{C}\bm{\Sigma}\bm{C}^T+\bm{R}$ of the linearized observation, the training objective reads
\begin{align}
\mathcal{L}(\btheta;\, \bx) = -\frac{1}{2} \sum_{k=1}^{N_k} \Bigg[ & \log\!\left|\bm{C}_{k|N_k}\, \bm{\Sigma}_{k|N_k}\, \bm{C}_{k|N_k}^T + \bm{R}\right| \nonumber\\
& + \left(\bx_k - h_{\phi}(\bmu_{k|N_k})\right)^T \left(\bm{C}_{k|N_k}\, \bm{\Sigma}_{k|N_k}\, \bm{C}_{k|N_k}^T + \bm{R}\right)^{-1} \left(\bx_k - h_{\phi}(\bmu_{k|N_k})\right) \nonumber\\
& + n_x \log(2\pi) + \log\frac{\left|\bm{A}_{k-1|N_k}\, \bm{\Sigma}_{k-1|N_k}\, \bm{A}_{k-1|N_k}^T + \bm{Q}\right|}{\left|\bm{\Sigma}_{k|N_k}\right|} - n_z \nonumber\\
& + \mathrm{Tr}\!\left(\left(\bm{A}_{k-1|N_k}\, \bm{\Sigma}_{k-1|N_k}\, \bm{A}_{k-1|N_k}^T + \bm{Q}\right)^{-1} \bm{\Sigma}_{k|N_k}\right) \nonumber\\
& + \left(f_{\nu}(\bmu_{k-1|N_k}) - \bmu_{k|N_k}\right)^T \left(\bm{A}_{k-1|N_k}\, \bm{\Sigma}_{k-1|N_k}\, \bm{A}_{k-1|N_k}^T + \bm{Q}\right)^{-1} \nonumber\\
& \quad \times \left(f_{\nu}(\bmu_{k-1|N_k}) - \bmu_{k|N_k}\right) \Bigg] \label{eq:loss}
\end{align}
 
In variational inference, the objective typically focuses on the reconstruction loss, which is dominated by the inference and observation models, while the transition function plays a comparatively minor role.
As a result, training can end up with an accurate inference model but a transition function that is too weak for reliable response prediction.
To address this, the Neural \gls{EKF} adopts \emph{replay overshooting}, originally proposed in~\cite{li2021replay}: an additional term that propagates the latent state forward through the generative model and penalizes the resulting prediction error.
This term enforces consistency between the transition model and the true latent dynamics over multiple steps, rather than only at the one-step level.
Specifically, a sequence of means and covariances $\left[\bmu_{k|N_k},\, \bm{\Sigma}_{k|N_k}\right]$ and the initial value $\left[\bar{\bmu}_0,\, \bar{\bm{\Sigma}}_0\right] = \left[\bmu_{0|N_k},\, \bm{\Sigma}_{0|N_k}\right]$ is used for prediction.
Analogously to the prediction stage of the \gls{KF}, the predicted values are obtained from the generative model as
\begin{gather}
\bar{\bmu}_k = f_{\nu}(\bar{\bmu}_{k-1}, \bu_k)\\
\bar{\bm{\Sigma}}_k = \bm{A}_{k-1}\, \bar{\bm{\Sigma}}_{k-1}\, \bm{A}_{k-1}^T + \bm{Q}
\end{gather}
which leads to a second set of distributions $\bar{q}_{\btheta}(\bz_k) = \mathcal{N}\!\left(\bar{\bmu}_k,\, \bar{\bm{\Sigma}}_k\right)$ obtained from the generative model.
The final reconstruction term in the loss is therefore composed of both posteriors, namely those obtained from \gls{EKF}-based inference and those obtained from the transition model.
These terms are weighted by a coefficient $\lambda_{\text{ro}}$ to form the training objective:
\begin{align}
\mathcal{L}(\btheta;\, \bx) = \sum_{k=1}^{N_k} \Bigg( & \lambda_{\text{ro}}\, \mathbb{E}_{q_{\btheta}(\bz_k)}\!\left[\log p_{\phi}(\bx_k|\bz_k)\right] + (1 - \lambda_{\text{ro}})\, \mathbb{E}_{\bar{q}_{\btheta}(\bz_k)}\!\left[\log p_{\phi}(\bx_k|\bz_k)\right] \nonumber\\
& - \mathbb{E}_{q_{\btheta}(\bz_{k-1})}\!\left[\mathrm{KL}\!\left(q_{\btheta}(\bz_k) \,\|\, p_{\nu}(\bz_k|\bz_{k-1}, \bu_k)\right)\right] \Bigg)
\label{eq:loss_ro}
\end{align}
where the second term is the addition relative to~\autoref{eq:elbo1} that accounts for overshooting.
 
\begin{practicaltip}
The gradual introduction of the replay-overshooting term in~\autoref{eq:loss_ro} requires careful scheduling to balance reconstruction accuracy and the learning of predictive dynamics. Introducing this term too rapidly may destabilize optimization, whereas delaying its contribution excessively may limit the training of the transition model and compromise long-horizon forecasting accuracy. A suitable strategy is to begin with a short burn-in phase, progressively increase the overshooting contribution, and employ gradient clipping and learning-rate decay to improve numerical stability. The duration of the ramping schedule and the associated optimization settings should be selected based on validation performance, considering both reconstruction and open-loop forecasting accuracy.
\end{practicaltip}
 
\subsubsection{Dynamic state estimation}\label{sec:NEKFresponse}
The formulation above does not, by itself, prescribe the physical meaning of the latent coordinates. For the dynamic state estimation problem considered here, we introduce an additional structural bias by constraining the Neural \gls{EKF} latent space to an interpretable nodal mechanical state,

\begin{equation}
\bz_k=[\bm{q}_k;\bm{v}_k]\in\mathbb{R}^{20},
\end{equation}
where $\bm{q}_k,\bm{v}_k\in\mathbb{R}^{10}$ denote displacement- and velocity-like coordinates at the monitored channels.
Rather than learning an unrestricted transition map, a shared acceleration network
$f_{\btheta}^{a}:\mathbb{R}^{20}\rightarrow\mathbb{R}^{10}$ is embedded in a fixed integrator of the second-order mechanical equations.
With a signed learnable input map $\bm{B}_u$, the acceleration and transition mean are
\begin{align}
\bm{a}_k &=
f_{\btheta}^{a}(\bm{q}_k,\bm{v}_k)
+\bm{B}_u\bu_k,\\
\bm{v}_{k+1} &= \bm{v}_k+\Delta t\,\bm{a}_k,\\
\bm{q}_{k+1} &= \bm{q}_k+\Delta t\,\bm{v}_{k+1},
\end{align}
which is the same semi-implicit (symplectic) Euler update adopted by the \gls{PgDMM} in~\autoref{sec:PgDMMresponse}, with the velocity advanced first and the updated velocity then advancing the displacement.
The difference with respect to the \gls{PgDMM} is that the acceleration field is learned here, so part of the first-order truncation error is absorbed into $f_{\btheta}^{a}$ during training.
The input map is initialized from the known state-space feed-through, and the acceleration field is decomposed into an explicit linear structural backbone and a neural residual,
\begin{equation}
f_{\btheta}^{a}(\bm{q}_k,\bm{v}_k)
=
-\left(\bm{K}_\ell\bm{q}_k+\bm{C}_\ell\bm{v}_k\right)
+
f_{\btheta}^{\mathrm{res}}(\bm{q}_k,\bm{v}_k),
\label{eq:nekf_backbone}
\end{equation}
with $\bm{K}_\ell,\bm{C}_\ell$ the stiffness and damping of the linear backbone, obtained by least squares on the training window and thereafter optimized jointly with the residual.
Unlike the \gls{PgDMM} in~\autoref{sec:PgDMMresponse}, whose operators come from the reduced linear model and are kept fixed, the backbone here is identified from data.
The linear backbone provides a warm start; if omitted, the network learns the entire state-to-acceleration mapping.

\begin{shadednote}
The numerical stability of the semi-implicit Euler integration scheme imposes a time-step-dependent restriction on the frequencies that can be accurately represented. For the undamped linear backbone in~\autoref{eq:nekf_backbone}, stability requires $\omega \Delta t < 2$, corresponding to an upper frequency limit of approximately $32$~Hz for $\Delta t=0.01$~s. Modes exceeding this threshold may render the discrete transition unstable, even when the identified continuous-time dynamics are physically admissible. Introducing additional numerical damping may suppress the resulting instability but can distort the identified dynamics, leading to artificially rapid decay of the predicted response. This illustrates an important distinction between discretization-induced errors and deficiencies in the identified model. In particular, accurate measurement-conditioned reconstruction does not guarantee faithful autonomous dynamics, as such discrepancies may only become apparent through open-loop forecasting. Where necessary, smaller integration steps, substepping, or more suitable time-integration schemes should be considered.
\end{shadednote}

The observation consists of the ten accelerations and reuses the same acceleration network and input map as the transition.
Owing to this sharing, the parameter partition $\btheta = [\nu;\, \phi]$ collapses in this specialization, and both models are subscripted by $\btheta$ directly.
Let $\bm{b}_a$ denote the learned acceleration bias and
$\bm{\tau}=\bm{\sigma}_{\mathrm{disp}}\oslash\bm{\sigma}_{\mathrm{acc}}$ the channel-wise ratio of a displacement to an acceleration standard deviation over the training windows, held fixed.
The observation mean is
\begin{equation}
h_{\btheta}(\bz_k,\bu_k)=
\bm{\tau}\odot
\bm{a}(\bm{q}_k,\bm{v}_k,\bu_k)+\bm{b}_a .
\label{eq:nekf_response_emission}
\end{equation}
The emitted acceleration is therefore, by construction, the second derivative of the latent displacement, and since no output scale is learned, the latent state carries displacement and velocity units although neither quantity is measured.
What an acceleration record cannot fix is the level of these states: integrating it twice determines the displacement only up to an offset and a linear drift, $\bm{d}_0+\bm{v}_0 t$, and the velocity only up to an offset, and nothing in the learned acceleration field forces these components to be resolved.
The state-recovery comparison of~\autoref{sec:responseOnlyStates} is accordingly scored with them removed.
The displacement scale in $\bm{\tau}$ and the least-squares fit of the backbone use displacements obtained by integrating the measured training accelerations, so no displacement record enters the model.

The displacement process variance is fixed near zero to preserve the kinematic identity, whereas velocity process variance, measurement variance, and the initial Gaussian state are learned.

The \gls{EKF} prediction and measurement updates of~\autoref{sec:KF}, followed by Rauch-Tung-Striebel smoothing, provide the Gaussian posterior without a separate inference network.
For this specialization, the objective combines the smoothed reconstruction negative log-likelihood $\mathcal{L}_{s}$, the filtering (innovation) marginal negative log-likelihood $\mathcal{L}_{f}$, a one-step dynamics \gls{KL} term, and an open-loop Gaussian prediction loss $\mathcal{L}_{\mathrm{pred}}$ over the window suffix:
\begin{equation}
\mathcal{J}_{\mathrm{NEKF}}
=
\lambda_{\mathrm{ro}}\mathcal{L}_{s}
+
\lambda_{f}\mathcal{L}_{f}
+
\lambda_{\mathrm{KL}}\mathcal{L}_{\mathrm{KL}}
+
(1-\lambda_{\mathrm{ro}})\lambda_{\mathrm{pred}}
\mathcal{L}_{\mathrm{pred}}.
\label{eq:nekf_response_loss}
\end{equation}
Here $\lambda_{\mathrm{ro}}$ balances reconstruction against prediction as in the replay-overshooting objective above, and $\lambda_{f}$, $\lambda_{\mathrm{KL}}$ and $\lambda_{\mathrm{pred}}$ weight the filtering, \gls{KL} and prediction terms.
The filtering term
\begin{equation}
\mathcal{L}_{f}
=
-\sum_{k=1}^{N_k}\log\mathcal{N}\!\left(\bx_k;\, h_{\btheta}(\bmu_{k|k-1},\bu_k),\, \bm{S}_k\right),
\qquad
\bm{S}_k=\bm{C}_k\bm{\Sigma}_{k|k-1}\bm{C}_k^T+\bm{R},
\end{equation}
is computed from the innovation sequence already produced by the filtering pass, so it costs nothing beyond the recursions of~\autoref{sec:KF}.
It scores one-step-ahead predictive accuracy under the model's own uncertainty and is therefore the dynamics-faithful counterpart to the smoothed term $\mathcal{L}_{s}$, which sees every measurement.
The prediction term propagates the filtered belief at the end of an $N_w$-step prefix over the remainder of the training window and scores the suffix, i.e., the same open-loop forecast used for evaluation.

\begin{practicaltip}
The prediction-loss weight $\lambda_{\mathrm{pred}}$ in~\autoref{eq:nekf_response_loss} directly influences the accuracy of open-loop forecasting and therefore requires careful calibration. Selecting this hyperparameter based on a single validation window may lead to unreliable model rankings, particularly when predictive performance varies across excitation realizations or initial conditions. A more robust strategy involves evaluating candidate configurations across multiple validation windows and random seeds, while reserving an independent test set for final performance assessment. The differences observed between validation and test performance in~\autoref{sec:responseOnlyTraining} further illustrate the sensitivity of model selection to the evaluation data. For the numerical experiments presented in~\autoref{sec:responseOnly}, $\lambda_{\mathrm{pred}}=5$ is adopted.
\end{practicaltip}

\subsubsection{Joint dynamic state and parameter estimation}\label{sec:NEKFparam}
 
Analogously to the \gls{PgDMM} extension in~\autoref{sec:PgDMMparam}, the Neural \gls{EKF} can be augmented to recover physical parameters by appending a dedicated parameter block to the latent state, $\bz = \left[\bz^{\text{dyn}},\, \bz^{\text{par}}\right]$.
The dynamic block retains the learnable transition $f_\nu$ and the observation map $h_\phi$, while the parameter block evolves under an identity mean transition with a small process-noise covariance term that injects parameter uncertainty into the EKF prediction step.
Within the \gls{EKF}, the parameter block contributes additional rows and columns to the Jacobian matrices $\bm{A}_k, \bm{C}_k$ and to the corresponding covariances $\bm{\Sigma}_{k|k}$, allowing the filter to propagate parameter uncertainty alongside the dynamic state.
A supervised loss term on the inferred parameter block is added to the \gls{ELBO} of~\autoref{eq:elbo1}, mirroring the construction described for the \gls{PgDMM}, and accordingly enabling joint state and parameter estimation within a single end-to-end training procedure.
The closed-form inference inherited from the \gls{EKF} is naturally compatible with this extension, since the augmented state remains a Gaussian random variable.

\subsection{Reduced-order modeling for a dynamics-informed latent space}\label{sec:ROMback}

The \gls{DMM} and Neural \gls{EKF} introduced above illustrate two levels of structure within deep Bayesian state-space modeling. The \gls{DMM} learns an unrestricted latent representation, whereas the structural specialization of the Neural \gls{EKF} organizes the latent state into displacement- and velocity-like coordinates and constrains their evolution through mechanical kinematics. A further level of physics-based inductive bias can be introduced by informing not only the organization of the state space, but also its governing dynamics and observation mappings through a compact model derived from prior physical knowledge. For structural systems, \glspl{ROM} provide a particularly useful mechanism for this purpose: they approximate the dominant dynamics of a high-dimensional system within a low-dimensional subspace while retaining a direct connection to physically meaningful states and observation channels. Embedding such a representation within a deep Bayesian formulation can reduce the effective dimension of the inference problem, constrain the latent dynamics, and facilitate interpretation of the inferred state in terms of the underlying mechanical system. Physics-based \glspl{ROM} are typically obtained from a high-dimensional \gls{FE} representation by extracting a reduced subspace, for example via modal analysis, Proper Orthogonal Decomposition, or Krylov-based reduction~\cite{benner2017model, giacomini2026surrogates, quinn2012modal}.

Regardless of the reduction technique adopted, the resulting \gls{ROM} can be cast in continuous-time state-space form as
\begin{gather}
\dot{\bz}(t) = \bm{A}_c(\bp)\, \bz(t) + \bm{B}_c(\bp)\, \bu(t), \label{eq:rom_cont_state}\\
\bx(t) = \bm{C}_c(\bp)\, \bz(t) + \bm{D}_c(\bp)\, \bu(t), \label{eq:rom_cont_obs}
\end{gather}
where $\bz \in \mathbb{R}^{n_z}$ is the reduced state, $\bu$ is the input excitation, and $\bx$ is the measured response. 
The matrices $\bm{A}_c(\bp), \bm{B}_c(\bp)$ encode the reduced dynamics, while $\bm{C}_c(\bp), \bm{D}_c(\bp)$ are the observation maps that project the reduced state and input onto the physical measurement channels. 
The dependence on the parameter vector $\bp$ acknowledges that the \gls{ROM} operators vary with the system properties to be identified.
Discretizing the system in~\autoref{eq:rom_cont_state} and~\autoref{eq:rom_cont_obs} under a zero-order hold at the sampling interval yields the discrete-time form
\begin{gather}
\bz_k = \bm{A}_d(\bp)\, \bz_{k-1} + \bm{B}_d(\bp)\, \bu_k, \label{eq:rom_disc_state}\\
\bx_k = \bm{C}_d(\bp)\, \bz_k + \bm{D}_d(\bp)\, \bu_k, \label{eq:rom_disc_obs}
\end{gather}
which provides the matrices $\bm{A}_d(\bp), \bm{B}_d(\bp), \bm{C}_d(\bp), \bm{D}_d(\bp)$ used as the physics-based component of the frameworks developed in the next subsections. 
The rows of $\bm{C}_d(\bp)$ and $\bm{D}_d(\bp)$ retain a direct correspondence to specific physical \glspl{DOF} of the underlying \gls{FE} representation: different row selections can potentially recover different physical observation channels from the same reduced state. 

For nonlinear and parameter-dependent systems, the construction of a globally accurate \gls{ROM} over the full operating range is challenging, and parametric or local \glspl{ROM} are often adopted~\cite{goutaudier2023new, opreni2023high, rozza2024parametric, vlachas2024parametric}.
We here adopt the local-basis approach of~\cite{vlachas2025reduced}, in which a library of locally trained \glspl{ROM} is assembled across the parameter space and interpolated, so that the reduced representation adapts to the operating regime rather than remaining fixed.
Such a representation provides a compact physics-based approximation of the dominant dynamics, while unresolved nonlinearities and model discrepancy can subsequently be delegated to learnable components.

Within deep Bayesian filtering, \glspl{ROM} therefore offer one mechanism for introducing physical structure into the latent representation, transition dynamics, and observation map without requiring the entire model to be prescribed.
The \gls{PgDMM} considered next provides a direct realization of this idea within the temporal variational framework, combining a physics-guided generative component with learnable terms.

\subsection{The Physics-guided Deep Markov Model (\gls{PgDMM})}\label{sec:PgDMM}

Building on the data-driven \gls{DMM} of~\autoref{sec:DMM} and the physics-informed latent representations discussed in~\autoref{sec:ROMback}, the \gls{PgDMM} provides a hybrid realization of the temporal variational state-space framework.
Rather than leaving the latent dynamics entirely unconstrained, available physical knowledge is introduced into the generative model, while learnable components account for dynamics and discrepancies not captured by the prescribed physics.
Importantly, the physical component need not be restricted to a \gls{ROM}; a \gls{ROM} provides the particular structured representation adopted in the applications considered here.

The \gls{PgDMM} extends the \gls{DMM} baseline of~\autoref{sec:DMM} along three lines:
\begin{enumerate}[label=(\roman*)]
    \item it retains the \gls{DMM} variational inference network to obtain the latent state $\bz$ from the observations $\bx$;
    \item it couples a deep generative model with a physics-based state-space model in the transition, with the two contributions combined additively;
    \item it learns the variance terms as functions of the latent state, allowing the framework to capture uncertainty arising both from the model and from the measurements.
\end{enumerate}
By incorporating prior knowledge of the model structure into the generative path, the \gls{PgDMM} encourages a latent-space representation that aligns with the expected physics.
At the same time, a flexible \gls{NN} models the discrepancy between the physics-based component and the monitored system to account for missing knowledge.
The resulting hybrid framework yields an inferred latent space that is interpretable from a physical standpoint while still capturing uncertainties related to the model and the measurements.
The \gls{PgDMM} further inherits the marginal-likelihood decomposition in~\autoref{eq:prob_DMM} and the Markov factorization in Equations~\ref{eq:transitionDMM} and~\ref{eq:observationDMM} from the \gls{DMM}, and modifies only the parameterization of the transition and observation densities to include a physics-guided component.

\subsubsection{Transition model}

The \gls{PgDMM} casts the transition from $\bz_{k-1}$ to $\bz_k$ as a hybrid structure comprising a physics-guided module $f_{\text{phy}}$ and a learning-based module $f_l$ parameterized by \glspl{NN}.
The updated latent state $\bz_k$ is computed as a linear weighted combination of two random variables $\bz_k^{\text{phy}}$ and $\bz_k^{\text{NN}}$:
\begin{gather}
\bz_k = \lambda\, \bz_k^{\text{phy}} + (1 - \lambda)\, \bz_k^{\text{NN}} \label{eq:DMMtransit} \\
\bmu(\bz_k) = \lambda\, \bmu_{\text{phy}}(\bz_{k-1}^{\text{phy}}) + (1 - \lambda)\, \bmu_{\text{NN}}(\bz_{k-1}^{\text{NN}}), \\
\bm{\Sigma}(\bz_k) = \lambda^2\, \bm{\Sigma}_{\text{phy}}(\bz_{k-1}^{\text{phy}}) + (1 - \lambda)^2\, \bm{\Sigma}_{\text{NN}}(\bz_{k-1}^{\text{NN}})
\end{gather}
where $\lambda$ is a trainable mixing parameter, i.e., it is included in the vector $\btheta$.
The physics-guided random variable $\bz^{\text{phy}}$ follows a Gaussian distribution whose mean is updated according to $f_{\text{phy}}$ and whose variance is emitted by a trainable \gls{NN}:
\begin{gather}
\bz_k^{\text{phy}} \sim \mathcal{N}\!\left(\bmu_{\text{phy}}(\bz_{k-1}^{\text{phy}}),\, \bm{\Sigma}_{\text{phy}}(\bz_{k-1}^{\text{phy}})\right) \label{eq:zphy} \\
\bmu_{\text{phy}}(\bz_{k-1}^{\text{phy}}) = f_{\text{phy}}(\bz_{k-1}^{\text{phy}}) \label{eq:meanz}\\
\bm{\Sigma}_{\text{phy}}(\bz_{k-1}^{\text{phy}}) = g_{\sigma}(\bz_{k-1}^{\text{phy}}) \label{eq:varz}
\end{gather}
where $g_{\sigma}$ is composed of feed-forward \glspl{NN}.
Following the same construction for $\bz^{\text{NN}}$, and mirroring the physics branch:
\begin{gather}
\bz_k^{\text{NN}} \sim \mathcal{N}\!\left(\bmu_{\text{NN}}(\bz_{k-1}^{\text{NN}}),\, \bm{\Sigma}_{\text{NN}}(\bz_{k-1}^{\text{NN}})\right)\\
\bmu_{\text{NN}}(\bz_{k-1}^{\text{NN}}) = f_l(\bz_{k-1}^{\text{NN}}), \qquad
\bm{\Sigma}_{\text{NN}}(\bz_{k-1}^{\text{NN}}) = g_{z}(\bz_{k-1}^{\text{NN}}) \label{eq:znn}
\end{gather}
where the learnable module $f_l$ and the variance network $g_{z}$ are both composed of feed-forward \glspl{NN}.

The transition function $f_{\text{phy}}$ for the random variable $\bz^{\text{phy}}$ has a physical interpretation and is formulated based on pre-existing knowledge of the system response, or by adopting a structured model for part of the transition process.
A typical example for nonlinear dynamical systems is the use of a linearized approximation of the dynamics around an operating point.
In turn, the random variable $\bz^{\text{NN}}$ captures the discrepancy between the approximation offered by $\bz^{\text{phy}}$ and the observed dynamics, via the learnable \glspl{NN}.
Because the combined latent state is anchored by $f_{\text{phy}}$ when maximizing the likelihood, training is less prone to learning arbitrary functions that simply fit the observations.
The covariance matrices $\bm{\Sigma}_{\star}$ are typically assumed to be diagonal, which is a sufficient approximation in most cases when uncertainties are adequately described by a white-noise assumption with a low noise-to-signal ratio.

\begin{practicaltip}
When the physics-guided transition is constructed from a \gls{ROM}, the dimension of its state representation must be consistent with that of the physics-guided latent subspace $\bz^{\text{phy}} \in \mathbb{R}^{n_z^{\text{phy}}}$. Since the operators defining $f_{\text{phy}}$ act directly on these latent coordinates, dimensional incompatibility can invalidate the resulting state-space operations. It is therefore advisable to verify the dimensions of the reduced operators and their compatibility with the latent-state representation during \gls{ROM} construction, prior to model training.
\end{practicaltip}

\subsubsection{Observation model}

The observation process that recovers the observations $\bx$ from the latent state $\bz$ is formulated as
\begin{equation}
\bx_k \sim \mathcal{P}\left(E(\bz_k)\right)
\end{equation}
where $\mathcal{P}$ denotes the probability distribution of the measured data $\bx$, parameterized by a suitable \gls{NN}.
For instance, Gaussian distributions can be parameterized as in~\autoref{eq:znn} via their mean and variance, with $E(\bz_k) = \left[\bmu(\bz_k),\, \bm{\Sigma}(\bz_k)\right]$, the mean $\bmu(\bz_k)$ being produced by the observation function $h_\phi$ detailed below; other distributions can be treated accordingly.
For the structural-response setting considered in this work, the observation function $h_\phi$ admits the additive decomposition
\begin{equation}\label{eq:emission_decomp}
\hat{\bx}_k = \underbrace{\bm{C}_d(\bp)\, \bz_k^{\text{phy}} + \bm{D}_d(\bp)\, \bu_k}_{\text{physics-derived emission}} + \underbrace{h_\phi^{\text{res}}(\bz_k)}_{\text{neural residual}},
\end{equation}
where $\bm{C}_d(\bp)$ and $\bm{D}_d(\bp)$ are the \gls{ROM} readout matrices that map the reduced physical state and the input to the measured \glspl{DOF}, and $h_\phi^{\text{res}}$ is a learnable head that absorbs the discrepancy between the \gls{ROM} output and the measurements.

\begin{practicaltip}
Consistent normalization is essential when combining physics-based and \gls{NN}-based components within the \gls{PgDMM}. While normalization can improve neural network training, the \gls{ROM}-derived transition and observation operators must remain consistent with the scaling applied to the latent states, inputs, and outputs. Inconsistent transformations can distort the relative contributions of the physics-guided and learned components, compromising the fidelity of the resulting dynamics without necessarily producing numerical errors. Scaling transformations should therefore be applied consistently across the state-space representation, with particular attention to the physical units and normalization conventions of the \gls{ROM} operators.
\end{practicaltip}

\subsubsection{Inference model}\label{sec:PgDMMinference}

The true posterior distribution $p(\bz_{1:N_k}|\bx_{1:N_k})$ associated with~\autoref{eq:prob_DMM} is intractable in the \gls{DMM} formulation.
It is approximated by a distribution $q_\rho(\bz_{1:N_k}|\bx_{1:N_k}) \approx p(\bz_{1:N_k}|\bx_{1:N_k})$ parameterized by $\rho$, which serves as the inference model that recovers the latent states from the observations.
The inference network employed here follows the standard structure of the \gls{DMM}.
Following the conditional independence statements in Equations~\ref{eq:transitionDMM} and~\ref{eq:observationDMM}, the latent variables $\bz_{1:N_k}$ are assumed to retain Markovian behavior when conditioned on $\bx_{1:N_k}$, and the true posterior is factorized as
\begin{equation}
p(\bz_{1:N_k}|\bx_{1:N_k}) = p(\bz_0) \prod_{k=1}^{N_k} p(\bz_k|\bz_{k-1}, \bx_{1:N_k})
\end{equation}
where $\bz_0$ is a trainable pseudo-node representing the initial conditions.
The approximate distribution $q_{\rho}$ is designed as a Gaussian that combines the physics-guided stream controlled by $\bz^{\text{phy}}$ and the learning-based stream by $\bz^{\text{NN}}$:
\begin{gather}
q_\rho(\bz|\bx_{1:N_k}) = q_\rho^{\text{phy}}(\bz^{\text{phy}}|\bx_{1:N_k})\, q_\rho^{\text{NN}}(\bz^{\text{NN}}|\bx_{1:N_k}) \nonumber \\
= q_\rho^{\text{phy}}(\bz_0^{\text{phy}})\, q_\rho^{\text{NN}}(\bz_0^{\text{NN}}) \prod_{k=1}^{N_k} q_\rho^{\text{phy}}(\bz_k^{\text{phy}}|\bz_{k-1}^{\text{phy}}, \bx_{1:N_k}) \prod_{k=1}^{N_k} q_\rho^{\text{NN}}(\bz_k^{\text{NN}}|\bz_{k-1}^{\text{NN}}, \bx_{1:N_k}) \label{eq:distributionq}\\
q_\rho^{\text{phy}}(\bz_k^{\text{phy}}|\bz_{k-1}^{\text{phy}}, \bx_{1:N_k}) = \mathcal{N}\!\left(\bmu_\rho^{\text{phy}}(\bz_{k-1}^{\text{phy}}, \bx_{1:N_k}),\, \bm{\Sigma}_\rho^{\text{phy}}(\bz_{k-1}^{\text{phy}}, \bx_{1:N_k})\right) \\
q_\rho^{\text{NN}}(\bz_k^{\text{NN}}|\bz_{k-1}^{\text{NN}}, \bx_{1:N_k}) = \mathcal{N}\!\left(\bmu_\rho^{\text{NN}}(\bz_{k-1}^{\text{NN}}, \bx_{1:N_k}),\, \bm{\Sigma}_\rho^{\text{NN}}(\bz_{k-1}^{\text{NN}}, \bx_{1:N_k})\right)
\end{gather}
Analogously to~\autoref{eq:znn}, the mean values and covariance matrices are obtained as outputs of dedicated \glspl{NN}:
\begin{gather}
\left[\bmu_\rho^{\text{phy}}(\bz_{k-1}^{\text{phy}}, \bx_{1:N_k}),\, \bm{\Sigma}_\rho^{\text{phy}}(\bz_{k-1}^{\text{phy}}, \bx_{1:N_k})\right] = g_\rho^{\text{phy}}(\by_k^{\text{phy}}) \\
\left[\bmu_\rho^{\text{NN}}(\bz_{k-1}^{\text{NN}}, \bx_{1:N_k}),\, \bm{\Sigma}_\rho^{\text{NN}}(\bz_{k-1}^{\text{NN}}, \bx_{1:N_k})\right] = g_\rho^{\text{NN}}(\by_k^{\text{NN}}) \\
\by_k^{\star} = \frac{1}{3}\left[\by_k^b + \by_k^f + \tanh\!\left(\bm{W}^{\star} \bz^{\star}_{k-1} + \bm{b}^{\star}\right)\right] \label{eq:yterm}
\end{gather}
In~\autoref{eq:yterm}, the superscript $\star$ denotes either the physics or the \gls{NN} stream, and the observation sequence $\bx_{1:N_k}$ is processed by a bidirectional recurrent \gls{NN} that outputs two sequences of hidden states $\by_k^f$ and $\by_k^b$, representing forward and backward information flow.
These past and future sequences are then combined with the latent state via a feed-forward network that transforms the latent state into the same dimension as the hidden sequences $\by_k^f$ and $\by_k^b$.
Although inference is conducted separately for the two streams, the same bidirectional network is applied, which implies that the parameters $\rho$ are shared.
The hidden states $\by_k^f$ and $\by_k^b$ are therefore identical for the two inference networks, and the distinction between the streams is encoded in the weights $\bm{W}^{\star}$ and biases $\bm{b}^{\star}$ in~\autoref{eq:yterm}.

\subsubsection{Evidence lower bound objective}\label{sec:ELBOpgdmm}

System identification and the learning of the corresponding dynamical models given a training dataset $\bx_{1:N_k}$ are formalized as identifying the vector of parameters $\btheta$ involved in the transition and observation models.
This is accomplished by maximizing the marginal log-likelihood function $\log p_{\btheta}\left(\bx_{1:N_k}\right)$, with $p_{\btheta}$ defined in~\autoref{eq:prob_DMM}, following the principle of maximum-likelihood estimation.
The true posterior over the latent variables given the observations is intractable, as explained in~\autoref{sec:PgDMMinference}, and variational inference is therefore used to derive the distribution $q_\rho$ in~\autoref{eq:distributionq} that is parameterized by \glspl{NN} and approximates the true posterior.
The \gls{ELBO} of the marginal log-likelihood $\log p_{\btheta}\left(\bx_{1:N_k}\right)$ is then adopted as the loss function to be maximized, after the temporal dependence and the factorization in Equations~\ref{eq:transitionDMM} and~\ref{eq:observationDMM} are accounted for, alongside the modeling assumption in~\autoref{eq:DMMtransit}.
A detailed derivation for the \gls{PgDMM} can be found in the original publication~\cite{liu2022physics}, and the resulting loss is reproduced here:
\begin{align}
\log p_{\btheta}(\bx) &\geq \mathcal{L}_{\text{phy}}(\btheta, \rho;\, \bx) \nonumber \\
&\!:= \mathbb{E}_{q_\rho}\!\left[\log p_{\btheta}(\bx|\bz)\right] - \left(\mathrm{KL}\!\left[q_\rho^{\text{phy}}(\bz^{\text{phy}}|\bx) \,\|\, p_{\btheta}(\bz^{\text{phy}})\right] + \mathrm{KL}\!\left[q_\rho^{\text{NN}}(\bz^{\text{NN}}|\bx) \,\|\, p_{\btheta}(\bz^{\text{NN}})\right]\right) \nonumber \\
&= \sum_{k=1}^{N_k} \mathbb{E}_{q_\rho}\!\left[\log p_{\btheta}(\bx_k|\bz_k)\right] - \sum_{k=1}^{N_k} \mathbb{E}_{q_\rho^{\text{phy}}(\bz_{k-1}^{\text{phy}}|\bx_{1:N_k})} \!\left[\mathrm{KL}\!\left(q_\rho^{\text{phy}}(\bz_k^{\text{phy}}|\bz_{k-1}^{\text{phy}}, \bx_{1:N_k}) \,\|\, p_{\btheta}(\bz_k^{\text{phy}}|\bz_{k-1}^{\text{phy}})\right)\right] \\
&\quad - \sum_{k=1}^{N_k} \mathbb{E}_{q_\rho^{\text{NN}}(\bz_{k-1}^{\text{NN}}|\bx_{1:N_k})} \!\left[\mathrm{KL}\!\left(q_\rho^{\text{NN}}(\bz_k^{\text{NN}}|\bz_{k-1}^{\text{NN}}, \bx_{1:N_k}) \,\|\, p_{\btheta}(\bz_k^{\text{NN}}|\bz_{k-1}^{\text{NN}})\right)\right] \nonumber
\end{align}
This loss encodes the trade-off between reconstruction accuracy, reflected by the expected log-likelihood term, and prior regularization, reflected by the \gls{KL}-divergence terms.
The physics-informed structure of the model splits the prior regularization into two terms, where one acts as a physical constraint.

\subsubsection{Dynamic state estimation}\label{sec:PgDMMresponse}

For the dynamic state estimation setting, a single-branch structural specialization is adopted rather than the two-stream mixture in~\autoref{eq:DMMtransit}.
The latent variable is the balanced reduced mechanical state
\begin{equation}
\bz_k = [\bm{r}_k;\bm{v}_k]\in\mathbb{R}^{32},
\end{equation}
comprising 16 relative modal displacements and 16 modal velocities.
Run-specific mass-normalized stiffness and damping operators are extracted from the linear reduced model and transformed using per-coordinate scales obtained from the discrete controllability Gramian.
This yields the balanced operators $\bm{K}_b$, $\bm{C}_b$, $\bm{g}_b$, the per-coordinate ratio $\bm{s}_b$ of the velocity to the displacement scale, the displacement and acceleration readouts $\bm{\Phi}_b^{r}$ and $\bm{\Phi}_b^{a}$, and the direct feed-through $\bm{D}_b$, i.e., the second-order, balanced counterparts of the operators in~\autoref{eq:rom_disc_state} and~\autoref{eq:rom_disc_obs}.
The only learned mean-dynamics component is a residual nonlinear restoring-force network
$f_{\btheta}^{\mathrm{res}}:\mathbb{R}^{32}\rightarrow\mathbb{R}^{16}$; there is no free neural transition branch and no mixture coefficient.
This network is shared between the transition and the emission below, so the parameter partition $\btheta = [\nu;\, \phi]$ collapses in this specialization and the generative components are subscripted by $\btheta$ directly.

Throughout this work, subscripts are used to distinguish the different roles of $\bm{C}$. Mechanical damping matrices are identified by their coordinate representation ($\bm{C}_b$ here and $\bm{C}_\ell$ in~\autoref{sec:NEKFresponse}), while the \gls{ROM} readout operators are distinguished by their continuous- or discrete-time formulation ($\bm{C}_c$ and $\bm{C}_d$, respectively). The observation Jacobian is denoted by $\bm{C}_k$, with the subscript indicating the time step, as introduced in~\autoref{sec:KF}. The stiffness matrix is consistently denoted by $\bm{K}$.

Each sampling interval $\Delta t=0.01$~s is divided into $n_s=10$ semi-implicit Euler substeps of size $\delta t=\Delta t/n_s$, since the retained modal spectrum, which reaches $\omega\approx970$~rad/s, violates the stability bound $\Delta t\,\omega<2$ of a single step per interval.
With $\bu_{k,j}$ the input sample of substep $j$, taken from the excitation record at its original $1$~kHz rate, the transition mean is evaluated at substep $j$ as
\begin{align}\label{eq:pgdmm_response_transition}
\bm{a}^{\mathrm{abs}}_j
    &= -\bm{K}_b\bm{r}_j-\bm{C}_b\bm{v}_j
       -f_{\btheta}^{\mathrm{res}}(\bm{r}_j,\bm{v}_j),\\
\bm{a}^{\mathrm{rel}}_j
    &= \bm{a}^{\mathrm{abs}}_j+\bm{g}_b\,\bu_{k,j},\\
\bm{v}_{j+1}
    &= \bm{v}_{j}+\delta t\,\bm{a}^{\mathrm{rel}}_j,\\
\bm{r}_{j+1}
    &= \bm{r}_{j}+\delta t\,
       \left(\bm{s}_b\odot\bm{v}_{j+1}\right).
\end{align}
The velocity is updated before displacement, giving a semi-implicit mechanical integrator.
After the final substep, the transition density is diagonal Gaussian,
\begin{equation}
p_{\btheta}(\bz_k|\bz_{k-1},\bu_k)
=
\mathcal{N}\!\left(f_{\btheta}(\bz_{k-1},\bu_k),
\operatorname{diag}(\bm{\sigma}_w^2)\right),
\end{equation}
with learned process scales $\bm{\sigma}_w$, i.e., $\bm{Q}=\operatorname{diag}(\bm{\sigma}_w^{2})$.

For the dynamic state estimation setting, the observations comprise accelerations measured at the monitored channels. These are modeled through the normalized acceleration projection $\bm{\Phi}_b^{a}$, while the emission model shares the residual restoring-force network employed in the transition model:
\begin{equation}
h_{\btheta}(\bz_k,\bu_k)
=
\bm{\Phi}_b^{a}
\left[
-\bm{K}_b\bm{r}_k-\bm{C}_b\bm{v}_k
-f_{\btheta}^{\mathrm{res}}(\bm{r}_k,\bm{v}_k)
\right]
+\bm{D}_b\,\bu_k
+\bm{b}_0 .
\label{eq:pgdmm_response_emission}
\end{equation}
Here $\bm{D}_b\bu_k$ is the direct feed-through of the reduced model, so that without the residual~\autoref{eq:pgdmm_response_emission} reproduces its acceleration output $\bm{C}_d\bz_k+\bm{D}_d\bu_k$, and $\bm{b}_0=-\bm{\mu}_x\oslash\bm{\sigma}_x$ is the fixed offset of the channel-wise normalization of the observations.
Neither displacement nor velocity is observed, so recovering either is a genuine test of the latent-state virtual sensing.
Unlike the Neural \gls{EKF} of~\autoref{sec:NEKFresponse}, the modal latent cannot drift freely, since any change of $\bm{r}_k$ passes through the fixed stiffness $\bm{K}_b$ into the emitted acceleration, where it is penalized by the measurements.
The observation density is Gaussian with learned diagonal measurement covariance $\bm{R}$.

Inference remains amortized but uses a single posterior stream.
An initialization encoder processes the first observation--input pairs, a bidirectional recurrent network summarizes the full sequence, and a Gaussian combiner merges this context with the preceding latent state and current input:
\begin{equation}
q_{\rho}(\bz_k|\bz_{k-1},\bx_{1:N_k},\bu_{1:N_k})
=
\mathcal{N}\!\left(\bmu_{\rho,k},
\operatorname{diag}(\bm{\sigma}_{\rho,k}^{2})\right).
\end{equation}
Training minimizes the negative \gls{ELBO} augmented by an open-loop forecast auxiliary,
\begin{equation}
\mathcal{J}_{\mathrm{PgDMM}} =
\underbrace{-\sum_{k=1}^{N_k}\mathbb{E}_{q_{\rho}}
\!\left[\log p_{\btheta}(\bx_k|\bz_k)\right]
+
\lambda_{\mathrm{KL}}
\sum_{k=1}^{N_k}
\mathrm{KL}\!\left[
q_{\rho}(\bz_k|\cdot)
\,\|\,p_{\btheta}(\bz_k|\bz_{k-1},\bu_k)
\right]}_{\text{negative \gls{ELBO}}}
+
\lambda_{\mathrm{fc}}\,\mathcal{L}_{\mathrm{fc}},
\label{eq:pgdmm_response_loss}
\end{equation}
where $\lambda_{\mathrm{KL}}$ weights the \gls{KL} term.
The auxiliary term addresses the exposure bias of the one-step objective: the \gls{ELBO} above never asks the transition to run without measurement feedback, so a model that reconstructs a window well can still forecast it poorly.
Given a rollout origin $k_0$ and a horizon $N_h$, the guide is warmed up on the $N_w$ observed steps preceding the rollout and the transition is then run free over the horizon, giving
\begin{equation}
\mathcal{L}_{\mathrm{fc}}
=
\frac{1}{2\sigma_{\mathrm{ref}}^{2}}
\sum_{k=k_0+N_w}^{k_0+N_w+N_h}
\left\|
\bx_k - h_{\btheta}\!\left(\bz_k^{\mathrm{roll}},\bu_k\right)
\right\|^{2},
\label{eq:pgdmm_forecast_aux}
\end{equation}
with $\bz_k^{\mathrm{roll}}$ propagated by the transition alone and $\sigma_{\mathrm{ref}}$ a fixed reference scale that renders the term commensurate with the reconstruction likelihood.
Here $N_w$ denotes the measurement warm-up length of~\autoref{tab:response_only_config}; the origin $k_0$ is sampled within the training windows and the horizon $N_h$ is increased gradually during training.

\begin{practicaltip}
The selection of rollout initial conditions and prediction horizons in~\autoref{eq:pgdmm_forecast_aux} can have a greater influence on forecasting accuracy than the weighting coefficient $\lambda_{\mathrm{fc}}$. Fixing the rollout origin at the beginning of each training segment limits the diversity of trajectories encountered during optimization. In contrast, randomly sampling rollout origins within each segment exposes the model to a broader range of initial conditions, while progressively increasing the horizon $N_h$ facilitates the learning of longer-term dynamics. In the numerical experiments of~\autoref{sec:responseOnlySetup}, this combined strategy yielded one of the most substantial improvements in forecasting performance. Importantly, the rollout horizon should remain representative of the intended prediction horizon: excessively short rollouts may improve computational efficiency and increase the number of sampled initial conditions, but provide insufficient constraints on long-term predictive behavior. Rollout sampling and horizon scheduling should therefore be considered jointly when designing the training procedure.
\end{practicaltip}

\subsubsection{Joint dynamic state and parameter estimation}\label{sec:PgDMMparam}

The standard \gls{PgDMM} formulation infers the latent dynamics governing the system response, but does not explicitly recover physical parameters.
We here extend the \gls{PgDMM} to address this need by augmenting the latent state with a dedicated parameter block, $\bz = \left[\bz^{\text{dyn}},\, \bz^{\text{par}}\right]$, where $\bz^{\text{par}} \in \mathbb{R}^{n_p}$ is a latent-space representation of the physical parameter vector $\bp$ defined in~\autoref{sec:Problem_parameter}.
The dynamic block $\bz^{\text{dyn}}$ is assembled from the physics-guided and neural streams $\bz^{\text{phy}}$ and $\bz^{\text{NN}}$ of~\autoref{eq:DMMtransit} and retains that hybrid transition, while the parameter block evolves according to a near-identity transition
\begin{equation}
\bz^{\text{par}}_k = \bz^{\text{par}}_{k-1} + \bm{\eta}_k
\end{equation}
with $\bm{\eta}_k$ a low-variance Gaussian innovation, reflecting the assumption that the targeted parameters are constant over the observation window.
At training time, the parameter values are treated as observations at every step, contributing a weighted Gaussian log-likelihood term to the \gls{ELBO} of~\autoref{eq:elbogeneric} that balances reconstruction and parameter recovery.
At inference time, the same network recovers $\bz^{\text{par}}$ from the response signal alone, by means of the \gls{ELBO}-trained inference model.
The resulting framework therefore conducts joint state and parameter estimation within the same probabilistic backbone, while preserving the physics-guided structure of the original \gls{PgDMM}.

\section{Demonstration on a Benchmark Shear Frame With Hysteretic Links}\label{sec:Applications}

Nonlinear system identification using the three frameworks of \autoref{sec:Methodology} is exemplified in a structural dynamics case study involving hysteretic nonlinearities.
Specifically, we consider a \gls{FE} model of a three-dimensional two-story two-bay shear frame with nonlinear joints, each exhibiting a Bouc-Wen type hysteretic behavior~\cite{carboni2016nonlinear}.
This example is a modified version of the modular, nonlinear multi-degree of freedom simulator that has been published as an open-access benchmark in~\cite{vlachas2021two}\footnote[3]{\url{https://github.com/KosVla/NonlinearBoucWenFrameBenchmark}}.
This demonstrative example is chosen due to the relevance of the simulator to real-life structural infrastructure assets and its inherent ability to parametrically model multiple simultaneously activated instances of nonlinearity, representing damage or condition deterioration in different regions, as exemplified in~\cite{vlachas2025reduced, zhang2026reduced, zhu2026discovering}.
The model is described next.

\subsection{Model description}

\begin{figure*}[!ht]
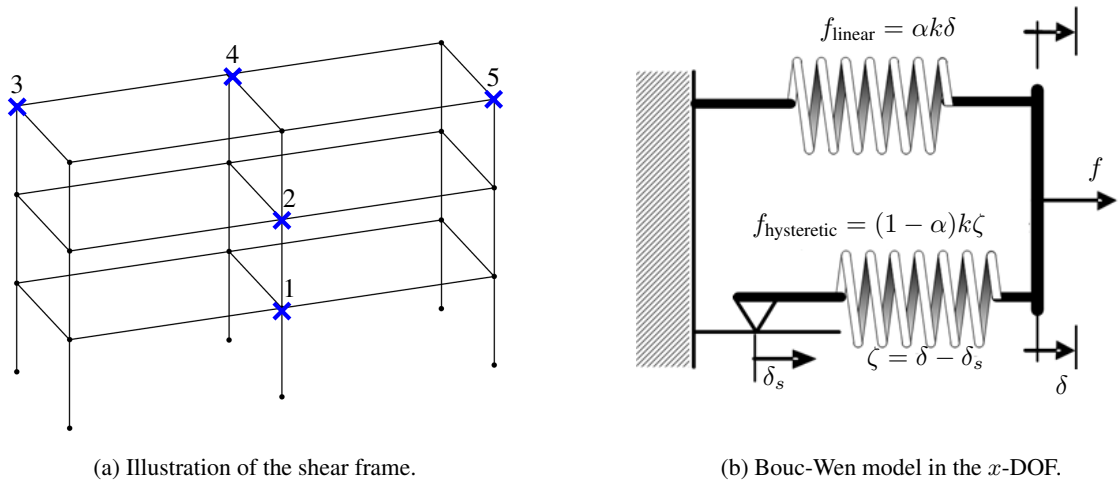

    \centering
    \begin{subfigure}[b]{0.48\textwidth}
    \centering
     \input{figures/Frame.tikz}
    \caption{Illustration of the shear frame. \label{fig:links}}
    \end{subfigure}
    \quad
    \begin{subfigure}[b]{0.48\textwidth}
     \centering
     \input{figures/BoucWenSpring.tikz}
     \caption{Bouc-Wen model in the $x$-DOF.\label{fig:bw}}
     \end{subfigure}
\caption{Schematic of the shear frame and link mechanism. Blue crosses denote sensor locations. \label{fig:BWmodel}}
\end{figure*}

A schematic of the considered shear frame is shown in~\autoref{fig:links}.
All joints in~\autoref{fig:links} are modeled using the Bouc-Wen hysteretic formulation, a smooth nonlinear model widely used to represent material hysteresis.
Specifically, a Bouc-Wen model is implemented at each \gls{DOF} of every nodal link to simulate the total restoring force $\bm{f}$.
An illustration of the nonlinear mechanism along the longitudinal $x$-\gls{DOF} is provided in~\autoref{fig:bw}.
In this formulation, the restoring force $\bm{f}$ is expressed as the superposition of linear and nonlinear terms, represented by the two springs in~\autoref{fig:bw}, and depends on the instantaneous nodal displacement $\bdelta$ and its hysteretic component $\bm{\zeta}$.
The vectorized governing equations for the restoring force of all \glspl{DOF} of a link are given as:
\begin{align}
\bm{f} = \bm{f}_{\text{linear}} + \bm{f}_{\text{hysteretic}} =  \alpha k \bdelta + (1-\alpha) k \bm{\zeta}
\label{eq:restforces}
\end{align}
where $\bdelta$ represents the nodal displacement, $\alpha$ denotes the characteristic post-yield to elastic stiffness ratio, and $k$ is the corresponding stiffness coefficient of the link.
The variable $\bm{\zeta}$ represents the hysteretic portion of the elongation, or deformation in general, and controls the hysteretic forcing.
It obeys the following
\begin{gather}
\dot{\bm{\zeta}} = A \dot{\bdelta} - \beta |{\dot{\bdelta}}|\bm{\zeta}|\bm{\zeta}|^{m-1}-\gamma {\dot{\bdelta}}|\bm{\zeta}|^{m}
\label{eq:boucwendegrad}
\end{gather}
where the shape, smoothness, and amplitude of the hysteretic curve that characterizes the dynamic behavior of each joint are determined by parameters $\beta$, $\gamma$, $m$, and $A$.
Their absolute values are not of interest, but rather their sums or differences, which may indicate a hardening or softening relationship~\cite{heredia2025state, ikhouane2007variation}.

Thus, the benchmark allows for a structural dynamics simulator, which is parameterized with respect to system properties and the traits of each joint's behavior.
Regarding material properties, steel HEA cross sections are used for all beam elements, and the structure is assembled using two parallel frames along the $x$-axis, spaced $5$~m apart, each spanning two bays of $7.5$~m.
All stories have a height of $3.2$~m.
The frame is further subjected to base excitation representing seismic loading.
Base accelerograms are applied at an angle of $\pi/4$ relative to the longitudinal axis.

\begin{table}[!hbt]
\caption{Modeling assumptions for the shear-frame. The uniform distribution in $(\star,\bullet)$ is $\mathcal{U}(\star,\bullet)$. The range of $w_e$ results from the temperature law given in the text.} %
\label{tab:input}
\centering
\begin{tabular}{l c c c c c c c }
\toprule
Parameter: & $w_e$ & $\alpha$ & $w_{a}$ & $\beta$ & $\gamma$  & $A,m$ & $w_k$\\
\midrule
Range: & $[0.74,1.32]$ & $0.2$ & $\mathcal{U}(0.0,1.0)$ & $3.0$ & $2.0$ & $1.0$ & $\mathcal{U}(0.1,1.0)$  \\
\bottomrule
\end{tabular}
\end{table}

The frameworks presented in~\autoref{sec:Methodology} are tasked to identify the underlying system and its response evolution in the presence of environmental and damage-induced variability.
The latter is simulated via varying the Young's modulus $E$ of the structure and the Bouc-Wen parameter $k$ in~\autoref{eq:restforces}.
Specifically, $E=w_e \cdot E_{\text{ref}}$ with $E_{\text{ref}}=210$~GPa, where $w_e = 0.15^{\,0.005\,(T - T_{\text{ref}})}$ models a temperature-dependent modulus driven by a seasonal temperature history $T$ with $T_{\text{ref}}=15^{\circ}$C, giving $w_e \in [0.74, 1.32]$, and $k=w_k \cdot k_{\text{ref}}, w_k \sim \mathcal{U}(0.10,1.0)$ and $k_{\text{ref}}=10^8$.
Regarding the excitation input, synthetic accelerograms are generated as weighted combinations of two reference seismic records:
$w_{a} \cdot (\text{El Centro}) + (1 - w_{a}) \cdot (\text{Morgan})$,
with the amplitude $w_{a} \sim \mathcal{U}(0.0, 1.0)$.
The modeling assumptions regarding the system's parametric variability are summarized in~\autoref{tab:input}.
The monitoring scheme retains five sensors (blue crosses in~\autoref{fig:links}), each recording acceleration in the longitudinal $x$ and transverse $y$ directions, and in both settings considered below the structural response is observed through these accelerations only.
This yields ten monitored channels, indexed once and used by every per-channel plot of this section: sensor by sensor following the numbering of~\autoref{fig:links}, longitudinal before transverse, so that odd indices are $x$-direction and even indices $y$-direction measurements.
Channel~4 in the figures below, for instance, is the transverse channel of the second sensor, denoted N2y where the per-channel labels are spelled out.
Simulated signals are polluted with additive white noise whose standard deviation is $5\%$ of the root mean square of each channel.

\subsection{Dynamic state estimation}\label{sec:responseOnly}
This setting addresses the dynamic state estimation problem formulated in~\autoref{sec:Problem_response}, where the physical system parameters remain fixed and only the evolving dynamic states are explicitly inferred.
No physical parameter is appended to the observation or represented in the latent state.
The three evaluations of~\autoref{sec:Problem_evaluation} are reported separately, instantiated on this task as follows:
\begin{itemize}
    \item \emph{Posterior reconstruction} uses the complete observation window through the \gls{PgDMM} variational guide or the Neural \gls{EKF} filter and smoother.
    \item \emph{Open-loop forecasting} conditions on only the warm-up prefix in~\autoref{tab:response_only_config} and propagates the remaining response under the known excitation without further measurement correction.
    \item \emph{Structural state recovery} maps the inferred modal or nodal displacement--velocity state back to physical coordinates and compares it with the hidden simulation state. Neither quantity is measured, and an acceleration record leaves a linear trend in displacement and a constant in velocity undetermined; before scoring, the recovered and the reference signals are therefore detrended alike, removing the mean from velocity and all content below $1$~Hz, which includes the linear trend, from displacement (\autoref{sec:responseOnlyStates}).
\end{itemize}
This scoring protocol separates these metrics to avoid labeling a full-sequence smoothed reconstruction as a forecast.
The complete set of numbers is collected in~\autoref{tab:response_only_metrics}.

\subsubsection{Setup}\label{sec:responseOnlySetup}

For each selected simulation run, the measured ground excitation and structural response are downsampled from $1$~kHz to $100$~Hz by stride sampling, divided into non-overlapping temporal windows, and split chronologically into training, validation, and test portions.
Channel-wise response and input normalization statistics are computed exclusively from the training windows and reused for validation and testing.
The resulting protocol evaluates later-time response modeling for the same physical system rather than generalization to unseen parameter values.
The dynamic state estimation setups are summarized in~\autoref{tab:response_only_config}.
They serve to demonstrate the two frameworks rather than constitute a controlled head-to-head comparison.
Beyond the latent parameterization, the two frameworks are supplied with different amounts of prior physics, which should be kept in mind when reading the results that follow.
The \gls{PgDMM} receives the modal operators of the reduced linear model, namely the modal stiffness, damping and base forcing of~\autoref{sec:PgDMMresponse}, as \emph{fixed} quantities that it never has to identify, and integrates them with ten symplectic substeps per observation interval.
The Neural \gls{EKF} receives only the known input feed-through to initialize its input gain, identifies its own linear backbone by least squares on the training window, and integrates with a single step per interval.
Both asymmetries favor the \gls{PgDMM} on the free-response behavior examined in~\autoref{sec:responseOnlyResults}.

\begin{table}[!ht]
\centering
\caption{Configurations of the dynamic state estimation experiments. The split reports the number of non-overlapping train/validation/test windows.}
\label{tab:response_only_config}
\resizebox{\textwidth}{!}{
\begin{tabular}{lcccccc}
\toprule
Method & run & observations & latent state & window & split & forecast warm-up \\
\midrule
\gls{PgDMM} & 131 & 10 accelerations & 16 modal displacements + 16 velocities & 200 steps (2 s) & 5/1/1 & 20 steps (0.2 s) \\
Neural \gls{EKF} & 131 & 10 accelerations & 10 nodal displacements + 10 velocities & 200 steps (2 s) & 5/1/1 & 20 steps (0.2 s) \\
\bottomrule
\end{tabular}}
\end{table}

The two frameworks are trained with the objectives of~\autoref{eq:pgdmm_response_loss} and~\autoref{eq:nekf_response_loss}, and for both the checkpoint with the lowest validation open-loop forecast \gls{RMSE} is retained, since the forecast is the reported quantity.

\subsubsection{Reconstruction and open-loop forecasting}\label{sec:responseOnlyResults}

\begin{figure}[!b]
\centering
\includegraphics[scale=0.99]{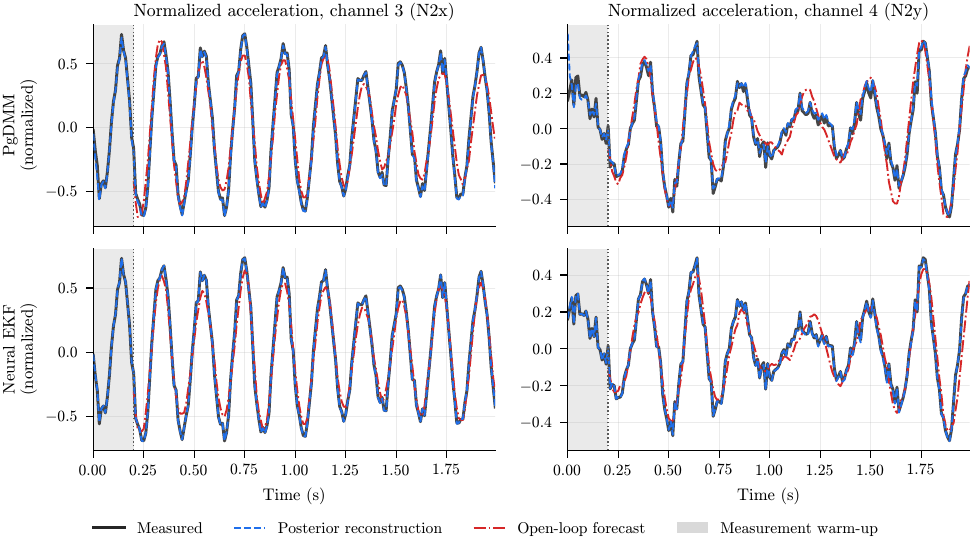}
\caption{Posterior reconstruction and open-loop forecasting on the held-out test window, for the dynamic state estimation \gls{PgDMM} (top) and Neural \gls{EKF} (bottom). The shaded interval is the $0.2$-s measurement warm-up. Left: normalized acceleration at monitored channel~3, which under the numbering of~\autoref{sec:Applications} is the longitudinal ($x$) channel of the second sensor in~\autoref{fig:links}; right: channel~4, its transverse ($y$) channel.}
\label{fig:response_only_forecast}
\end{figure}

Two time histories are reported in~\autoref{fig:response_only_forecast}, which contrasts, on the same held-out window, the posterior reconstruction against the open-loop forecast for the two measurement directions of the second sensor.
The corresponding error measures are summarized in~\autoref{tab:response_only_metrics}.

\begin{table}[!hbt]
\centering
\caption{Dynamic state estimation results on the held-out test window, for the setup of~\autoref{tab:response_only_config}.
Response errors are \gls{RMSE} of the ten accelerations, normalized by the training statistics of each channel, in which units the measurement noise has a standard deviation of $0.042$.
The three shorter forecast horizons pool the number of rollout origins given in parentheses.
State recovery is the normalized \gls{RMSE} between the projected latent state and the hidden simulation state, averaged over the ten monitored \glspl{DOF}; both projections are at raw scale, with nothing fitted to the reference.
The first two state-recovery rows score detrended signals, with all displacement content below $1$~Hz, which includes the linear trend an acceleration record cannot determine, and the velocity mean removed from the recovered and the reference signal alike; the next two score the projection as it stands.
The noise floor is the error of an estimate that matched the noise-free state exactly.}
\label{tab:response_only_metrics}
\begin{tabular}{lcc}
\toprule
 & \gls{PgDMM} & Neural \gls{EKF} \\
\midrule
\multicolumn{3}{l}{\emph{Posterior reconstruction} (complete observation window used)}\\
\quad Normalized \gls{RMSE}$_{\bx}$ & 0.0203 & 0.0386 \\
\midrule
\multicolumn{3}{l}{\emph{Open-loop forecast} (warm-up prefix only, then free-running dynamics)}\\
\quad Norm. \gls{RMSE}$_{\bx}$ over $0.2$~s horizon (17 origins) & 0.0675 & 0.0488 \\
\quad Norm. \gls{RMSE}$_{\bx}$ over $0.5$~s horizon (14 origins) & 0.0694 & 0.0539 \\
\quad Norm. \gls{RMSE}$_{\bx}$ over $1.0$~s horizon (9 origins) & 0.0789 & 0.0657 \\
\quad Norm. \gls{RMSE}$_{\bx}$ over full suffix ($1.8$~s, 1 origin) & 0.1020 & 0.1088 \\
\midrule
\multicolumn{3}{l}{\emph{Structural state recovery} (neither quantity is observed)}\\
\quad displacements, detrended & 0.2515 & 0.4068 \\
\quad velocities, detrended & 0.2945 & 0.3412 \\
\quad displacements as projected & 0.2719 & 20.6 \\
\quad velocities as projected & 0.2980 & 0.6469 \\
\quad noise floor, displacements / velocities & \multicolumn{2}{c}{0.180 / 0.163} \\
\midrule
selected checkpoint & epoch $14{,}700$ & iteration $5{,}500$ \\
selection metric & \multicolumn{2}{c}{validation forecast \gls{RMSE}} \\
\bottomrule
\end{tabular}
\end{table}

Both frameworks reconstruct the measured accelerations closely once every measurement on the monitoring window is assimilated.
The corresponding normalized \gls{RMSE} are $0.020$ for the \gls{PgDMM} and $0.039$ for the Neural \gls{EKF}, both below the standard deviation of the measurement noise ($0.042$).
The accuracy gap widens substantially once measurement feedback is no longer considered and the open-loop forecast mode is exercised: over the $1.8$-s suffix the forecast \gls{RMSE} rises to $0.102$ and $0.109$ respectively, five and almost three times the corresponding reconstruction error.
\begin{observation}
The substantial discrepancy between posterior reconstruction and open-loop forecasting, observed for both frameworks, highlights an important limitation of evaluating learned dynamical models solely through reconstruction accuracy. By assimilating measurements throughout the observation window, posterior inference can compensate for inaccuracies in the learned transition dynamics, potentially masking deficiencies that become apparent during autonomous prediction. In the present experiments, reconstruction errors underestimate open-loop forecasting errors by a factor of approximately three to five. These findings demonstrate that accurate measurement-conditioned reconstruction does not necessarily imply reliable predictive dynamics and underscore the importance of evaluating both capabilities separately through clearly defined assessment protocols.
\end{observation}

\begin{figure}[!ht]
\centering
\begin{subfigure}[b]{0.48\textwidth}
\centering
\includegraphics[width=\linewidth]{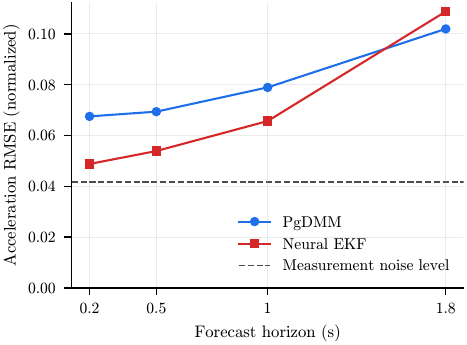}
\caption{Open-loop forecast \gls{RMSE} against the prediction horizon.\label{fig:response_only_horizon}}
\end{subfigure}
\hfill
\begin{subfigure}[b]{0.48\textwidth}
\centering
\includegraphics[width=\linewidth]{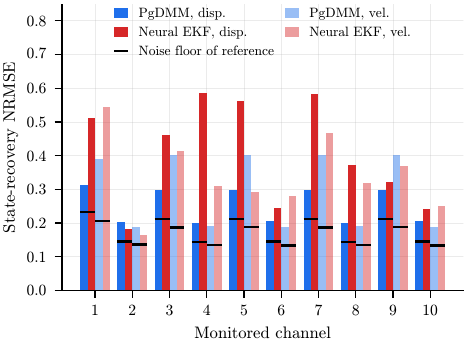}
\caption{Hidden-state recovery per monitored channel.\label{fig:response_only_recovery}}
\end{subfigure}
\caption{Horizon dependence of the open-loop forecast and per-channel hidden-state recovery, both on the held-out test window. The dashed line in~\autoref{fig:response_only_horizon} is the standard deviation of the measurement noise. State recovery in~\autoref{fig:response_only_recovery} is scored as in~\autoref{tab:response_only_metrics}, with displacement and velocity detrended, and the black segments mark the noise floor of the reference state. The alternating pattern reflects the two measurement directions: the odd- and even-indexed channels form two groups that are mutually correlated within a group and essentially uncorrelated across groups.}
\label{fig:response_only_panels}
\end{figure}

The dependence of the open-loop forecast error on the prediction horizon is also reported in~\autoref{tab:response_only_metrics} and~\autoref{fig:response_only_horizon}.
The Neural \gls{EKF} is the more accurate model over the shorter horizons, with $0.049$ against $0.068$ at $0.2$~s and $0.066$ against $0.079$ at $1.0$~s, but its error grows faster with the horizon, and over the full $1.8$-s suffix the two frameworks are on par ($0.102$ for the \gls{PgDMM} against $0.109$).
They also err in different directions: the \gls{PgDMM} forecast is least accurate in the longitudinal direction ($0.119$ against $0.081$ in the transverse one), the Neural \gls{EKF} in the transverse one ($0.124$ against $0.091$).
As~\autoref{fig:response_only_forecast} shows, both forecasts keep the phase of the response but lose amplitude in the longitudinal direction.
\begin{observation}
Embedding prescribed linear operators in the transition model can reduce the accumulation of forecasting errors, but does not necessarily ensure the physical fidelity of the predicted response. In the present experiments, the \gls{PgDMM}, which retains the modal stiffness and damping operators of the reduced linear model, exhibits a $1.5$-fold increase in forecast error between the shortest and longest prediction horizons, compared with a $2.2$-fold increase for the Neural \gls{EKF}, whose acceleration field is learned from data. Nevertheless, linearization of the trained \gls{PgDMM} transition reveals that the learned residual introduces additional effective damping in the dominant modes, with estimated damping ratios of $1$--$3\%$ for the three lowest modes, compared with $0.1$--$0.2\%$ in the underlying reduced model. This discrepancy results in an excessively rapid decay of the predicted response, illustrating that physics-guided representations can constrain forecast-error growth without fully preserving the system's underlying dynamics.
\end{observation}

\subsubsection{Recovery of the hidden structural state}\label{sec:responseOnlyStates}

\begin{figure}[!ht]
\centering
\includegraphics[width=\textwidth]{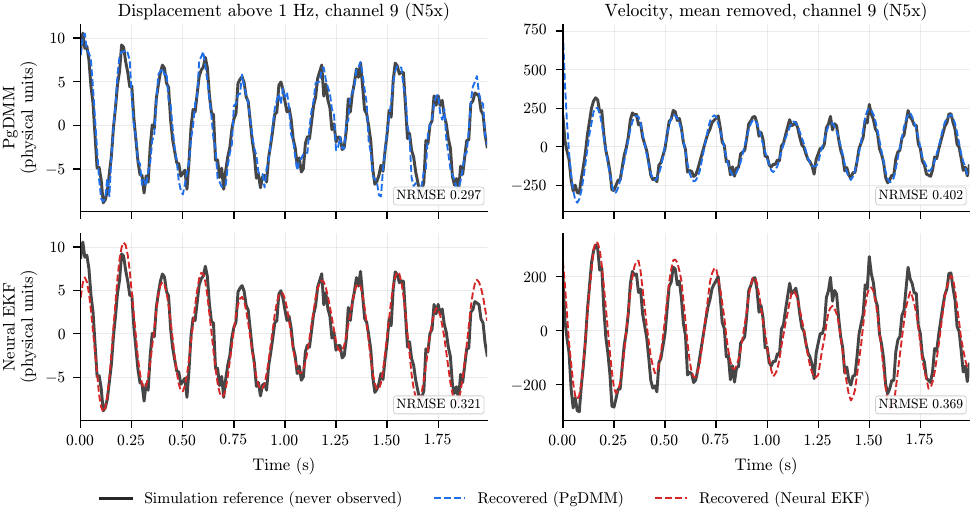}
\caption{Recovery of unmeasured structural displacement (left) and velocity (right) from the inferred latent states in the dynamic state estimation setting. Results are shown for channel~9, corresponding to the longitudinal ($x$) direction of the fifth sensor in~\autoref{fig:links}, selected because its estimation errors are closest to the average errors of each framework. Neither displacement nor velocity is directly observed by either model. The inferred states are projected at their original scale and processed consistently with the reference signals. For displacement, frequencies below $1$~Hz are removed, eliminating the integration-related linear trend that cannot be determined from acceleration measurements and accounting for no more than approximately $3\%$ of the reference displacement energy. For velocity, the mean is removed.}
\label{fig:response_only_states}
\end{figure}

The interpretability claim made for both frameworks is that their latent coordinates maintain a direct structured relationship with the physical displacements and velocities rather than posing as ad-hoc, arbitrary features.
This is tested directly in~\autoref{fig:response_only_states}, which projects each latent state back to the physical space and overlays the recovered quantities with the hidden full-order simulation state.
With acceleration measured alone, neither displacement nor velocity is observed, so both are substantive checks.
Both projections are made at raw scale: the Neural \gls{EKF} latent carries physical units through its fixed scale ratio and the balanced modal frame of the \gls{PgDMM} through the reduced model, so nothing is fitted to the reference state.

An acceleration record nevertheless leaves part of the state undetermined, a linear trend in displacement and a constant in velocity (\autoref{sec:NEKFresponse}), and the two frameworks treat these components very differently.
The \gls{PgDMM} latent does not drift, since its fixed stiffness turns any change of the modal displacement into an acceleration (\autoref{sec:PgDMMresponse}): detrending changes its errors by less than $10\%$, from $0.27$ to $0.25$ for displacement and from $0.30$ to $0.29$ for velocity.
The Neural \gls{EKF} displacement, by contrast, wanders far from the reference, with a normalized \gls{RMSE} of $20.6$ as projected.
Its error is concentrated at low frequency, where displacement produces almost no acceleration, whereas the oscillation above $1$~Hz is recovered with an error of $0.41$.
Linearizing the trained transition of the Neural \gls{EKF} exposes the mechanism, namely slow modes between $0.01$ and $1.3$~Hz, well below the first natural frequency of $3.9$~Hz, along which the displacement can drift with almost no effect on the emitted acceleration.

Displacement recovery is therefore evaluated above $1$~Hz in~\autoref{tab:response_only_metrics}, \autoref{fig:response_only_recovery} and~\autoref{fig:response_only_states}, a frequency band containing at least $97\%$ of the reference displacement energy across all channels.

Under this evaluation protocol, both frameworks recover the unobserved structural states, with the \gls{PgDMM} achieving greater accuracy, as shown in~\autoref{tab:response_only_metrics} and~\autoref{fig:response_only_recovery}. The \gls{PgDMM} yields normalized \gls{RMSE} values of $0.25$ for displacement and $0.29$ for velocity, with recovered amplitudes deviating by approximately $10\%$ or less from the reference across all channels. The corresponding errors for the Neural \gls{EKF} are $0.41$ and $0.34$, respectively. The \gls{PgDMM} also exhibits more uniform displacement recovery, with channel-wise errors ranging from $0.20$ to $0.31$, compared with $0.18$ to $0.58$ for the Neural \gls{EKF}. For both frameworks, the longitudinal channels exhibit larger relative errors on average, consistent with their smaller displacement amplitudes, which reach only $30$--$60\%$ of those in the transverse directions over the test window.

These results should be interpreted relative to the reference-state noise floor, corresponding to normalized errors of $0.18$ for displacement and $0.16$ for velocity, which would be obtained by an estimate that exactly reproduces the noise-free structural state.

\begin{observation}
A physically structured latent representation does not, by itself, guarantee accurate recovery of unobserved structural states. Although both frameworks assign a displacement--velocity interpretation to their latent coordinates, the Neural \gls{EKF} admits spurious low-frequency modes that permit substantial displacement drift with little influence on the measured accelerations. By contrast, the prescribed stiffness operators of the \gls{PgDMM} constrain these poorly observable components, resulting in more stable and accurate state recovery. This highlights the distinction between imposing physical structure on a latent representation and ensuring that the corresponding states can be reliably inferred from available measurements. Physics-based constraints can therefore contribute not only to interpretability, but also to reducing ambiguities in the recovery of unmeasured dynamic states.
\end{observation}

\subsubsection{Training behavior}\label{sec:responseOnlyTraining}

\begin{figure}[!ht]
\centering
\includegraphics[width=\textwidth]{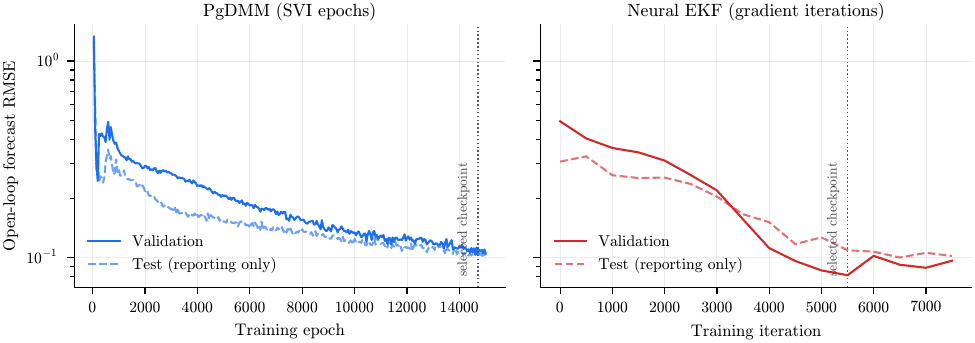}
\caption{Training convergence of the two frameworks in the dynamic state estimation setting, assessed through the open-loop forecast \gls{RMSE} on the validation (solid lines) and test (dashed lines) windows. The test data are used exclusively for independent performance assessment and do not inform model selection. The dotted vertical lines indicate the checkpoints selected according to each framework's validation criterion.}
\label{fig:response_only_convergence}
\end{figure}

The convergence behavior of both frameworks is reported in~\autoref{fig:response_only_convergence}, with the validation and test forecast errors tracked throughout training and the selected checkpoints marked.
The \gls{PgDMM} converges slowly: its validation error declines steadily over the whole budget, to $0.18$ at epoch $6{,}000$, $0.13$ at epoch $10{,}000$ and $0.10$ at the selected epoch $14{,}700$, and is still decreasing at the end of the $15{,}000$ epochs.
The Neural \gls{EKF} reaches $0.081$ at iteration $5{,}500$ and then oscillates between $0.089$ and $0.102$ without further improvement.

The validation and test curves in~\autoref{fig:response_only_convergence} also make the caveat of~\autoref{sec:responseOnlySetup} concrete.
They are computed on one window each and do not move together.
The two windows even rank the frameworks differently over the full suffix, the Neural \gls{EKF} leading on validation ($0.081$ against $0.103$) and the \gls{PgDMM} on test ($0.102$ against $0.109$), so small differences read off a single window should not be treated as evidence.

\subsection{Joint dynamic state and parameter estimation}\label{sec:responseParam}

In the joint dynamic state and parameter estimation setting, the models are tasked with simultaneously inferring the evolving structural states and the underlying physical system parameters.
All methods introduced in~\autoref{sec:Methodology} are considered, following the evaluation protocol of~\autoref{sec:responseOnly}, with results reported for posterior reconstruction and parameter recovery.

The open-loop forecasting evaluation of~\autoref{sec:responseOnly} is not repeated here, as it addresses a distinct aspect of model performance.
In that setting, a short measurement-conditioned prefix initializes the latent state, after which the response is propagated without further measurement updates.
For joint dynamic state and parameter estimation, this procedure remains well defined for the structural response but requires a different interpretation for the physical parameters, which are assumed time-invariant.
Once measurement updates cease, the parameter estimates remain fixed at the values inferred from the initial observation window, while prediction errors in the structural response accumulate over the forecast horizon.
Consequently, open-loop response forecasting and parameter recovery would assess different inference capabilities under this protocol and are therefore treated separately.
Furthermore, the open-loop predictive performance of the \gls{PgDMM} and Neural \gls{EKF} is already examined in~\autoref{sec:responseOnly} for a fixed physical system.
Recursive filtering schemes that support sequential inference as new measurements become available are reviewed in~\cite{tatsis2022sequential}.

\subsubsection{Setup}

The parameter distributions summarized in~\autoref{tab:input} are sampled to generate 190 simulation runs for training, validation, and testing.
Each run provides ten acceleration channels recorded at the five monitored nodes, comprising longitudinal ($x$) and transverse ($y$) measurements per sensor, together with three parameters: the Bouc-Wen stiffness weighting factor $w_k$, the Young's modulus weighting factor $w_e$, and the input excitation amplitude $w_a$.
Each simulation is performed with a time step of $dt=0.001$~s over $15$~s, subsequently downsampled by a factor of $10$, and divided into $2$~s windows.
The 190 runs are partitioned by run, so that all windows of a run fall in the same subset, into 133 training, 28 validation and 29 test runs (70\%/15\%/15\%, i.e., 931/196/203 windows of 2~s).
The test runs are never seen during training or validation.
The resulting configuration is summarized in~\autoref{tab:response_param_config}.

\begin{table}[!ht]
\centering
\caption{Default configuration for the joint dynamic state and parameter estimation experiments. All frameworks
observe the same ten acceleration channels and three system parameters, sampled at
$1$~kHz and downsampled by a factor of ten. The latent state is reported with its composition given in parentheses.}
\label{tab:response_param_config}
\resizebox{\textwidth}{!}{
\begin{tabular}{lcccc}
\toprule
Method & observations & latent state & window & split (runs) \\
\midrule
\gls{DMM} & \multirow{3}{*}{10 accelerations + 3 parameters} & 35 (32 unstructured + 3 parameters) & \multirow{3}{*}{200 steps (2 s)} & \multirow{3}{*}{133/28/29} \\
\gls{PgDMM} & & 35 (16 reduced displacements + 16 velocities + 3 parameters) & & \\
Neural \gls{EKF} & & 35 (32 unstructured + 3 parameters) & & \\
\bottomrule
\end{tabular}}
\end{table}

\subsubsection{Response reconstruction}\label{sec:resultsResponse}

\begin{figure}[!hbt]
\centering
\includegraphics[width=\textwidth]{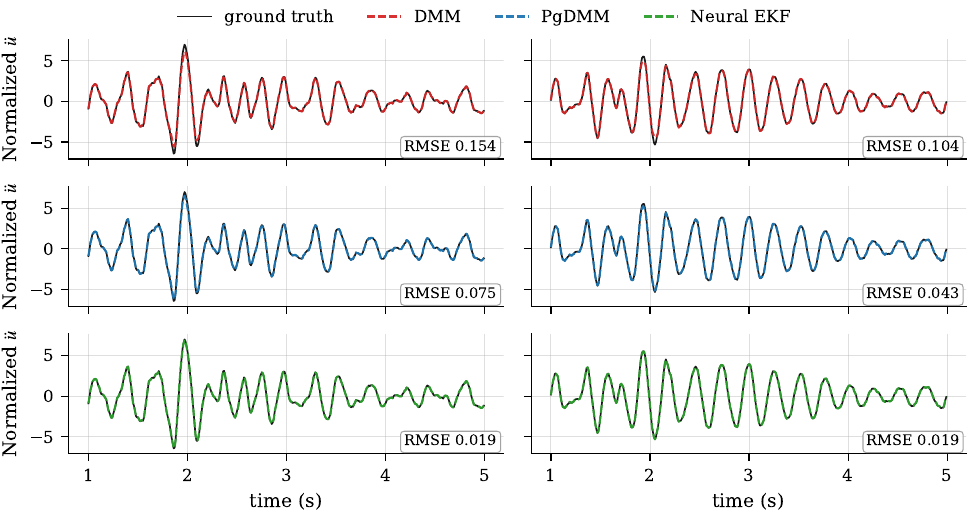}
\caption{Time-history response reconstruction on two representative test sequences, showing the reconstructed (dashed) against the true (solid) normalized acceleration $\ddot{u}$ on a monitored channel over the $1$--$5$\,s window. The two columns correspond to two different held-out runs, each reporting the transverse $y$-direction response at the top right monitored node of the second floor in~\autoref{fig:links}.}
\label{fig:response_test}
\end{figure}

The first evaluation concerns \emph{posterior response reconstruction}, following the scoring protocol of~\autoref{sec:responseOnly}.
A qualitative comparison of the three methods is presented in~\autoref{fig:response_test}, while the corresponding quantitative error measures are summarized in~\autoref{tab:metrics}.

\begin{table}[!hbt]
\centering
\caption{Joint dynamic state and parameter estimation on the held-out test windows and test records. The response entries refer to the monitored acceleration $\ddot{\bu}$ in normalized \gls{RMSE}. Every channel is standardized by its own mean and standard deviation and a value of $0.05$ corresponds to $5\%$ error of the (training) standard deviation of that channel.}
\label{tab:metrics}
\begin{tabular}{lccc}
\toprule
 & \gls{DMM} & \gls{PgDMM} & Neural \gls{EKF} \\
\midrule
\multicolumn{4}{l}{\emph{Response reconstruction} --- normalized acceleration $\ddot{\bu}$}\\
\quad \gls{RMSE} over the test windows & 0.0504 & 0.0336 & 0.0167 \\
\quad \gls{MAE} over the test windows & 0.0141 & 0.0111 & 0.0103 \\
\quad \gls{RMSE}, median over the entire test runs & 0.0262 & 0.0162 & 0.0142 \\
\midrule
\multicolumn{4}{l}{\emph{Parameter estimation} --- \gls{MAE} over the test windows, $\times 10^{-3}$}\\
\quad Bouc-Wen stiffness weighting $w_k$ & 1.71 & 1.44 & 2.17 \\
\quad Young's modulus weighting $w_e$ & 1.15 & 1.00 & 0.86 \\
\quad Excitation amplitude weighting $w_a$ & 1.91 & 1.60 & 0.12\\
\bottomrule
\end{tabular}
\end{table}

In~\autoref{fig:response_test}, the three methods are arranged in rows, while the columns correspond to two representative test sequences.
The reconstructed trajectories of the \gls{PgDMM} and Neural \gls{EKF} closely follow the reference responses, whereas the \gls{DMM} captures the dominant features but slightly underestimates the largest acceleration peaks.

The time histories are displayed over a longer continuous record for visualization, although training and quantitative evaluation are performed on the $2$~s windows specified in~\autoref{tab:response_param_config}.
Consecutive windows from the same simulation run are concatenated to reconstruct the continuous record, and inference is performed over the resulting sequence, with the $1$--$5$~s interval shown in the figure.
All three architectures can accommodate variable sequence lengths at inference, and the concatenated observations retain their original temporal ordering without introducing additional information beyond the measured response.

\begin{figure}[!ht]
\centering
\includegraphics[width=\textwidth]{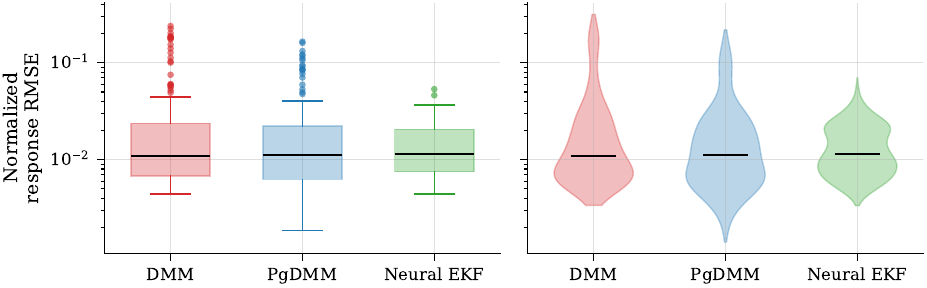}
\caption{Distribution of the normalized acceleration response \gls{RMSE} over the test windows shown as box plots (left) and as kernel density estimates (right). The vertical axis is logarithmic. The horizontal bar marks the median.}
\label{fig:F12violin}
\end{figure}

\begin{figure}[!ht]
\centering
\includegraphics[width=\textwidth]{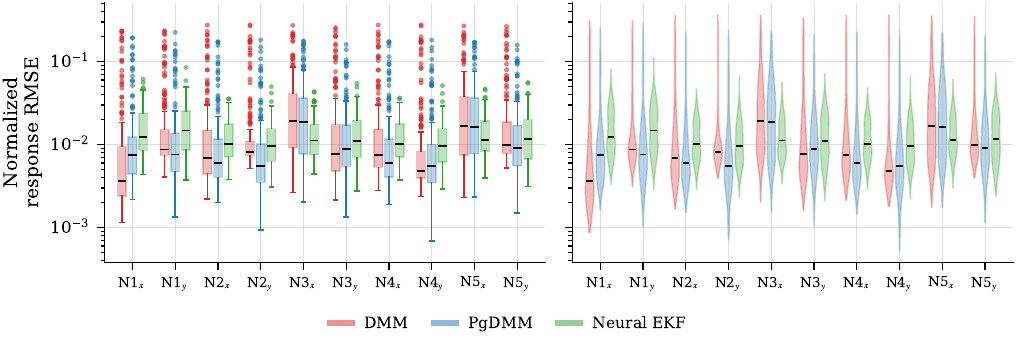}
\caption{Distribution of the normalized acceleration response \gls{RMSE} over the test windows per monitored channel shown as box plots (left) and as kernel density estimates (right). Channels are labeled by monitored node and direction following the numbering in~\autoref{fig:links}. The vertical axis is logarithmic.}
\label{fig:F-errbar}
\end{figure}

The distribution of reconstruction errors across the held-out test windows is presented in~\autoref{fig:F12violin} and summarized in~\autoref{tab:metrics}.
The corresponding spatial distribution across monitored channels is shown in~\autoref{fig:F-errbar}.
Both figures are constructed using the $2$~s testing windows.

All three frameworks achieve accurate posterior response reconstruction across the held-out test set, although systematic differences emerge when the full error distributions are considered.
As shown in~\autoref{fig:F12violin}, their typical performance is nearly indistinguishable, with per-window median normalized \gls{RMSE} values between $0.011$ and $0.012$.
More pronounced differences occur in the upper tails: the ninetieth-percentile error is $0.055$ for the \gls{DMM}, $0.038$ for the \gls{PgDMM}, and $0.025$ for the Neural \gls{EKF}.
Similarly, the proportions of test windows exceeding a normalized \gls{RMSE} of $0.05$ are $10.8\%$, $7.9\%$, and $0.5\%$, respectively.
The aggregate errors reported in~\autoref{tab:metrics} are therefore sensitive to the more challenging test realizations, revealing differences in performance consistency that are not apparent from the median values alone.

The channel-wise distributions in~\autoref{fig:F-errbar} indicate that these differences are not dominated by a single measurement location.
Although the error distributions are broadly comparable across channels, the longitudinal responses of the third and fifth monitored nodes exhibit the largest errors for the \gls{DMM} and the \gls{PgDMM}, whereas the largest Neural \gls{EKF} errors occur at the first monitored node.

\begin{observation}
Comparable median reconstruction errors do not necessarily imply equivalent robustness across operating conditions or parameter realizations.
Although all three frameworks achieve similar accuracy on typical test windows, their upper-tail errors differ substantially, with the Neural \gls{EKF} exhibiting the smallest proportion of high-error predictions.
This highlights the importance of evaluating the distribution of reconstruction errors, rather than relying exclusively on aggregate or median metrics, when assessing the reliability of learned state-space representations.
\end{observation}

\subsubsection{Parameter estimation}\label{sec:resultsParam}

The second evaluation concerns the recovery of physical system parameters from the observed structural responses.
Following the parameterization introduced in~\autoref{tab:input}, three quantities are inferred: the Bouc-Wen stiffness weighting factor $w_k$, the Young's modulus weighting factor $w_e$, and the excitation amplitude weighting factor $w_a$.
Each framework produces one estimate of each parameter per test window.

\begin{figure}[!hbt]
\centering
\includegraphics[width=\textwidth]{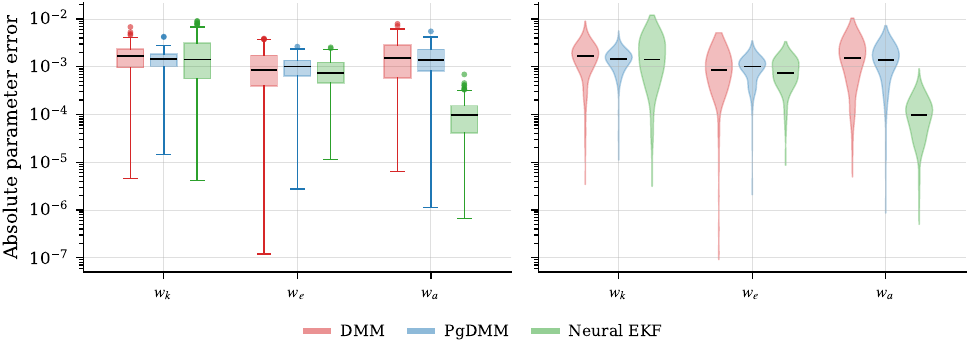}
\caption{Distribution of the absolute parameter error over the test windows shown as box plots (left) and as kernel density estimates (right). The vertical axis is logarithmic. The horizontal bar marks the median.}
\label{fig:param_error}
\end{figure}

The distributions of absolute estimation errors for the three weighting factors are presented in~\autoref{fig:param_error}, with the corresponding quantitative metrics summarized in~\autoref{tab:metrics}.
The \gls{PgDMM} estimates the parameters through the dedicated latent parameter block described in~\autoref{sec:PgDMMparam}, while the Neural \gls{EKF} employs the augmented-state construction of~\autoref{sec:NEKFparam}.
Both physics-enhanced frameworks generally outperform the \gls{DMM}, although the differences are relatively small.

\begin{figure}[!hbt]
\centering
\includegraphics[width=\textwidth]{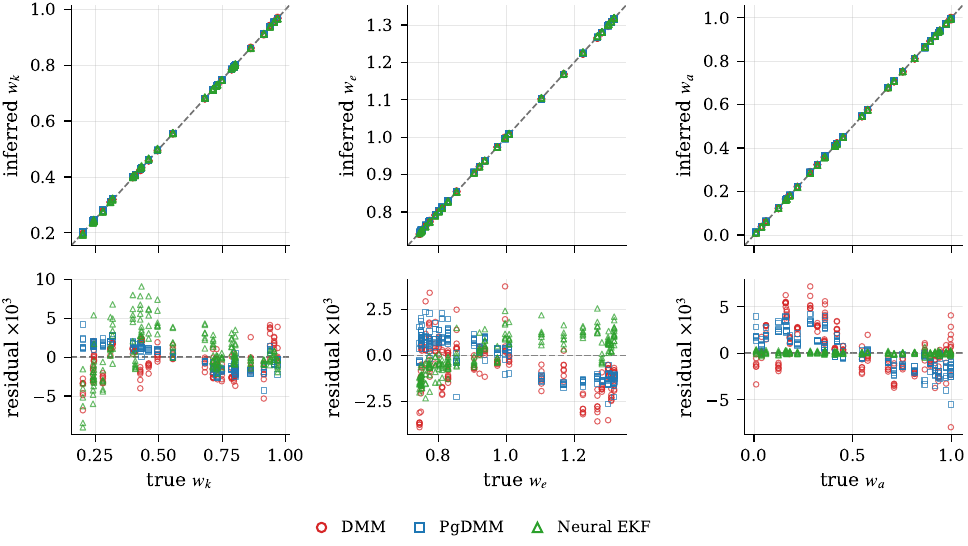}
\caption{Parameter estimation on the held-out testing windows. \emph{Top}: inferred against true values over all samples, with the dashed line marking exact recovery. \emph{Bottom}: the corresponding residuals, scaled by $10^{3}$.}
\label{fig:param_parity}
\end{figure}

The agreement between inferred and reference parameter values is examined in~\autoref{fig:param_parity}, which presents parity plots and estimation residuals across all testing windows.
For all three frameworks, the estimates closely follow the diagonal throughout the sampled range of each parameter, with the \gls{MAE} remaining below $1.0\%$ of the corresponding range in all nine model--parameter combinations.

Closer examination of the residuals in~\autoref{fig:param_parity} and the distributions in~\autoref{fig:param_error} reveals parameter-dependent differences.
The Young's modulus weighting factor $w_e$ is recovered with comparable accuracy by all three frameworks.
For the excitation amplitude weighting factor $w_a$, the Neural \gls{EKF} achieves approximately an order-of-magnitude improvement over the \gls{DMM} and the \gls{PgDMM}, the latter exhibiting an error of $1.6\times10^{-3}$.
For the Bouc-Wen stiffness weighting factor $w_k$, however, the Neural \gls{EKF} attains a median error comparable to those of the two Markov models but exhibits the largest mean error and the heaviest upper tail.
This distinction between typical and aggregate performance parallels the behavior observed for response reconstruction in~\autoref{sec:resultsResponse}.

The superior estimation accuracy of the Neural \gls{EKF} for $w_a$ may be attributed to the way excitation amplitude enters the model through the observed input, allowing measurement updates to directly inform its estimate during filtering.
By contrast, $w_k$ and $w_e$ influence the response through the latent structural dynamics, making their estimation more dependent on the adequacy of the learned transition model.
Across all three parameters, the \gls{PgDMM} improves upon the \gls{DMM}, consistent with the overall ordering of reconstruction performance reported in~\autoref{tab:metrics}.

\begin{observation}
Parameter recovery accuracy depends not only on the inference architecture but also on how individual parameters influence the measured response.
The Neural \gls{EKF} achieves particularly accurate estimation of the excitation amplitude $w_a$, whose influence enters through the observed input, while the estimation of parameters governing the latent structural dynamics exhibits greater variability.
The results indicate that parameter-specific error distributions, including their upper tails, are essential for assessing joint state and parameter estimation, since aggregate accuracy alone may conceal differences in the reliability of individual parameter estimates.
\end{observation}

\subsection{Interpretability of the inferred latent space}\label{sec:resultsInterp}

A central motivation for physics-aware deep Bayesian formulations is their potential to yield latent representations with a direct physical interpretation, in contrast to the unconstrained coordinates typically learned by purely data-driven models. We examine this property through the relative contributions of individual latent coordinates and their spectral correspondence with the underlying structural modes and input excitation. These complementary analyses assess whether the inferred dynamics retain a physically meaningful organization beyond their ability to reconstruct the measured response.

\begin{figure}[!ht]
\centering
\includegraphics[width=0.7\textwidth]{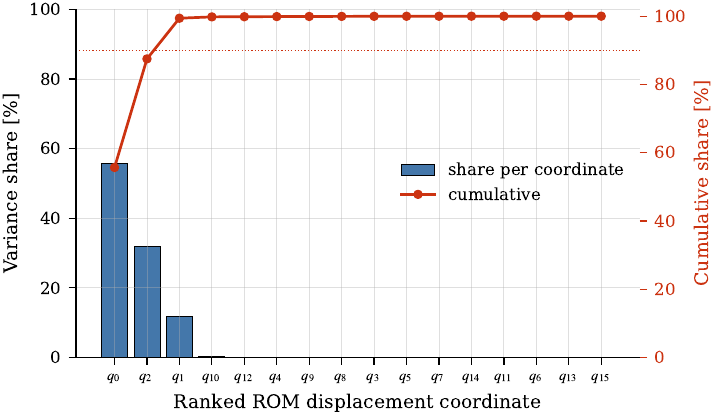}
\caption{Latent-coordinate variance ranking for the joint dynamic state and parameter estimation setting \gls{PgDMM} of~\autoref{sec:responseParam}, on a representative held-out testing run. The bars show the contribution of each of the sixteen latent displacement coordinates of the \gls{ROM} state to the total latent displacement variance. The cumulative contribution is also shown, with the $90\%$ level indicated.}
\label{fig:F7latent}
\end{figure}

The relative contributions of the latent coordinates are examined in~\autoref{fig:F7latent} for a representative held-out testing run of the \gls{PgDMM} in the joint dynamic state and parameter estimation setting. The first three modal displacement coordinates account for approximately $99\%$ of the total latent displacement variance, while the remaining coordinates contribute only marginally. This strong concentration indicates that, for the considered realization, the response is effectively represented by a small subset of the prescribed modal coordinates. It also illustrates how the physics-guided representation facilitates the identification of dominant contributions within the learned latent dynamics. These contributions are evaluated in terms of latent-coordinate variance and should not be interpreted directly as physical modal energies.

\begin{figure}[!ht]
\centering
\includegraphics[width=\textwidth]{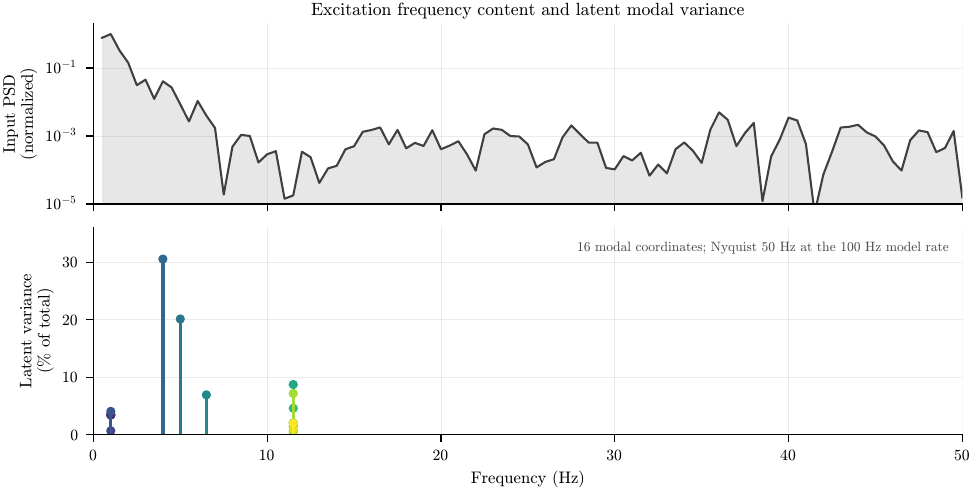}
\caption{Connection between the dominant latent modes and the frequency content of the input excitation, shown for the dynamic state estimation \gls{PgDMM} of~\autoref{sec:PgDMMresponse} on the held-out window of run~131. Top: power spectral density of the excitation, normalized to its peak. Bottom: variance of each of the modal displacement coordinates, as a share of their total variance, placed at the dominant frequency of its own latent trajectory and colored by increasing frequency.} %
\label{fig:F13modefreq}
\end{figure}

The correspondence between the latent dynamics and the modal characteristics of the underlying structure is further examined in~\autoref{fig:F13modefreq}. In structural dynamics, the response is often governed by a limited number of resonant modes, whose participation depends on both the excitation spectrum and the system's modal properties. A latent representation aligned with the physical modal decomposition therefore allows the inferred dynamics to be interpreted in terms of the structural modes contributing to the response.

For the dynamic state estimation \gls{PgDMM}, the first three modal coordinates correspond to the three lowest modes of the reduced model, with natural frequencies of $3.9$, $5.2$, and $6.3$~Hz. Their inferred trajectories exhibit dominant oscillations at these frequencies and jointly account for $58\%$ of the variance of the modal displacement coordinates. Eight additional coordinates exhibit dominant frequencies around $11.5$~Hz, close to the $11.1$--$11.3$~Hz modal cluster of the reduced model, and contribute a further $27\%$. The remaining five coordinates predominantly follow the lower-frequency excitation content, around $1$~Hz, consistent with the fact that $84\%$ of the excitation power lies below $2$~Hz. These variance contributions refer to the balanced coordinates introduced in~\autoref{sec:PgDMMresponse}, whose scaling depends on the controllability Gramian, and must therefore not be interpreted as physical modal energies.

\begin{figure}[!ht]
\centering
\includegraphics[width=\textwidth]{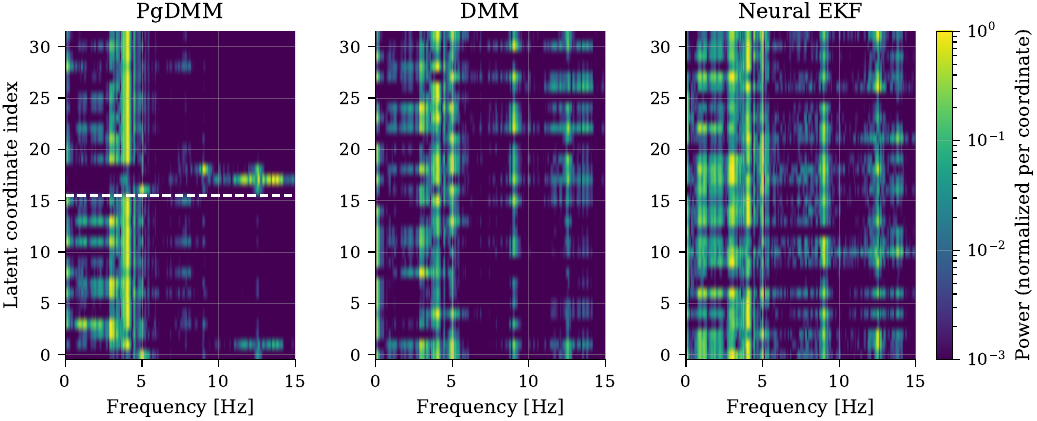}
\caption{Frequency content of the inferred latent space for the joint dynamic state and parameter estimation setting \gls{PgDMM}, \gls{DMM} and Neural \gls{EKF} of~\autoref{sec:responseParam}, on a representative held-out testing run. Each row shows the periodogram of a latent coordinate, normalized to its own peak and displayed on a logarithmic color scale. The horizontal axis is restricted to $15$~Hz, encompassing the principal excited frequency band. For the \gls{PgDMM}, the dashed line separates the displacement (bottom) and velocity (top) coordinates of its \gls{ROM} state. The corresponding coordinates of the other architectures are not organized according to the same prescribed modal partition.}
\label{fig:F18latentspectra}
\end{figure}

A complementary spectral analysis, extending to all three architectures in the joint dynamic state and parameter estimation setting of~\autoref{sec:responseParam}, is presented in~\autoref{fig:F18latentspectra}. The periodogram of each latent coordinate is normalized to its own peak, allowing the spectral organization of the representations to be compared independently of their relative amplitudes. Narrow-band concentration indicates that individual coordinates predominantly represent specific oscillatory components, whereas broader spectra indicate that several frequency contributions are mixed within the same coordinate.

For the \gls{PgDMM}, the latent coordinates exhibit predominantly narrow-band spectra concentrated around the lower resonances of the underlying linear \gls{ROM}, with little spectral content at higher frequencies in the displacement block. This organization is consistent with the prescribed modal structure and displacement--velocity partition inherited from the reduced model.

The \gls{DMM} and Neural \gls{EKF} also capture the principal frequency bands present in the measured response, but distribute these components across multiple latent coordinates without an evident correspondence between coordinate index and physical modal order. The Neural \gls{EKF} additionally exhibits substantial low-frequency content across several coordinates. 
In this setting its latent coordinates are unstructured and therefore not constrained to follow the prescribed modal basis of the \gls{PgDMM}. Consequently, the physics-guided formulation exhibits a more clearly organized modal representation, whereas the learned latent spaces exhibit greater spectral mixing, despite achieving comparable response-reconstruction accuracy on the considered realization.

\begin{observation}
Comparable response-reconstruction accuracy does not imply equivalent physical interpretability of the inferred latent states. The \gls{PgDMM} retains a prescribed modal organization, allowing individual coordinates to be related to structural modes and their participation under different excitations. By contrast, the more flexible learned representations distribute the same spectral content across multiple coordinates, making their physical interpretation less direct. This distinction highlights the role of physics-based inductive biases in promoting structured latent representations, beyond their contribution to predictive accuracy. Spectral alignment provides supporting evidence of such structure, although it does not by itself establish unique identification of the underlying physical modes.
\end{observation}

\section{Discussion}\label{sec:Discussion}
This tutorial has framed deep Bayesian filtering around three ingredients, namely a transition model, an observation model and a mechanism for posterior inference, each of which may be prescribed from physical knowledge, learned from data, or composed of both.
The \gls{DMM}, the \gls{PgDMM} and the Neural \gls{EKF} occupy different points of this design space, and the demonstrations of~\autoref{sec:Applications} illustrate how the choice made for each ingredient shapes what a trained model can deliver.
This section distills these observations into insights for the design of deep Bayesian filters, collects practical guidelines for their training and deployment, and outlines open directions for research.

\subsection{Key insights}

\paragraph{Physical structure constrains the latent representation.}
Learnable state-space models can reproduce the same measured response with many different latent representations; even in the linear case, input--output data determine a state-space model only up to a similarity transformation of its state.
Prescribed structure narrows this freedom.
Kinematic identities between displacement, velocity and acceleration fix the roles and units of the latent coordinates, and operators taken from a physics-based model fix how these coordinates interact, so that the latent state can be read in physical terms.
Structure also transfers the errors of the prescribed model, and the learned components that complement it can alter the prescribed dynamics, as the residual of the \gls{PgDMM} does when it adds damping to the dominant modes (\autoref{sec:responseOnlyResults}).
Physical knowledge should therefore be imposed where its validity is sufficiently established, while learned components should address unresolved dynamics or model discrepancies. Importantly, their influence on the prescribed physical behavior should be verified rather than assumed negligible.

\paragraph{The inference mechanism determines how a model can be deployed.}
Amortized inference networks, as used in the \gls{DMM} and the \gls{PgDMM}~\cite{krishnan2015deep, krishnan2017structured, karl2016deep}, require no linearization of the transition and observation models and scale readily to high-dimensional latent states, but they add a second learned model whose approximation error enters every posterior and which must itself generalize to unseen data.
When the inference network processes the sequence in both temporal directions, it returns smoothed rather than filtered estimates and cannot be used causally without modification.
Algorithmic inference, as in the Neural \gls{EKF}, instead derives the posterior from the learned model itself through the filtering recursions~\cite{haarnoja2016backprop, liu2024neural}.
It requires no separate inference-network parameters, is causal by construction, and yields innovation statistics that serve both likelihood evaluation and monitoring, at the price of the Gaussian and local-linearization assumptions of the underlying filter and of covariance computations whose cost grows cubically with the latent dimension.
Differentiable particle filters extend algorithmic inference beyond the Gaussian assumption~\cite{jonschkowski2018differentiable, karkus2018particle}, and filters with learned gains retain the recursive structure while learning part of the update~\cite{revach2022kalmannet}, so the inference mechanism, too, can be partly prescribed and partly learned.
The appropriate inference mechanism therefore depends on the intended application, the dimensionality of the latent state, and the required temporal processing. Recursive filtering provides a natural framework for causal online estimation, although covariance propagation can become expensive in high dimensions. Amortized inference offers greater flexibility in posterior approximation and may be advantageous for offline analysis or settings where inference costs can be reduced through learned mappings, provided that the inference architecture is compatible with the intended deployment mode.

\paragraph{The training objective determines what the model is good at.}
Variational objectives such as the \gls{ELBO} jointly constrain observation fidelity and latent transition dynamics, but do not necessarily enforce accurate autonomous predictions over extended horizons.
Posterior inference conditioned on measurements can compensate for local transition-model errors, allowing accurate reconstruction even when open-loop forecasting remains unreliable.
Additional predictive objectives, such as replay overshooting~\cite{li2021replay} and the open-loop prediction terms of~\autoref{sec:NEKFresponse} and~\autoref{sec:PgDMMresponse}, explicitly penalize accumulated prediction errors and can improve long-horizon fidelity.
Their effectiveness nevertheless depends on the prediction horizon considered during training, which should be representative of the intended deployment setting.
Parameter identification introduces a related challenge: the learning objective must also provide sufficient information to recover the physical quantities of interest.
Supervising a latent parameter block, as in the joint dynamic state and parameter estimation extensions of~\autoref{sec:NEKFparam} and~\autoref{sec:PgDMMparam}, requires labeled training configurations, which are readily available in simulation but may be difficult to obtain experimentally.
Alternatively, likelihood-based estimation using the marginal likelihood evaluated through a filtering recursion can support parameter inference without direct parameter labels~\cite{wigren2022nonlinear}, provided that the parameters are identifiable from the available observations.

\paragraph{Interpretability requires structure, observability and verification.}
A latent state becomes interpretable when its coordinates are tied to physical quantities by construction, such as the modal coordinates of a \gls{ROM} or the displacement--velocity partition of a mechanical state.
The physics-guided latent spaces of~\autoref{sec:resultsInterp} are organized in this way, with their dominant coordinates oscillating at the frequencies of the lowest structural modes, whereas unstructured latent states spread the same frequency content over many coordinates.
Construction alone does not suffice, however, since a coordinate is only as meaningful as the measurements allow it to be.
Where the data leave directions of the latent space undetermined, flexible components can place the estimate anywhere along them without degrading the reconstruction.
The double integration of acceleration records, which fixes displacement only up to an offset and a linear trend, is the structural example treated here, and learned transition models can additionally acquire spurious slow modes along which the latent state drifts with little effect on the observations (\autoref{sec:responseOnlyStates}).
Such directions should be identified, for instance through an observability analysis or an eigenanalysis of the linearized learned transition, and then be fixed by prior structure or additional sensing, or removed explicitly before latent states are compared with physical quantities.
That comparison, against states that were never measured, is what ultimately verifies an interpretation.
The same reasoning applies to physical parameters, whose estimates are meaningful only where the response is sensitive to them and no learned component can reproduce their effect.

\subsection{Practical guidelines for training and deployment}

\begin{itemize}
    \item \emph{Establish reference models.} A purely data-driven model such as the \gls{DMM} and the prescribed physics used on its own show what the learned and the prescribed components of a hybrid model contribute, and reveal when added structure does not pay off.
    \item \emph{Match the time discretization to the prescribed physics.} Operators derived from finite-element or reduced-order models often contain modes too stiff for an explicit update at the sampling interval, which calls for substepping or an exact discretization. Learned discrete-time maps absorb such effects but lose the direct link to continuous-time physics; continuous-time formulations~\cite{lai2021structural} offer an alternative.
    \item \emph{Keep scales consistent.} Normalizations applied to inputs, outputs and latent variables must be propagated to the prescribed operators, so that physics-based and learned components act on commensurate quantities.
    \item \emph{Train and select for the intended use.} Objectives should contain terms without measurement feedback whenever the model is to forecast, and checkpoints and hyperparameters should be chosen on the quantity that will be reported rather than on the training likelihood.
    \item \emph{Inspect the learned dynamics.} Linearization of the trained transition model, followed by comparison of its modal frequencies and damping characteristics with physical expectations, can reveal spurious, unstable, or excessively damped modes that may remain concealed by measurement-conditioned reconstruction.
    \item \emph{Evaluate separately, on several windows, and against the noise.} Reconstruction, forecasting, state recovery, and parameter recovery assess distinct model capabilities. In the demonstrations, open-loop forecast errors exceed reconstruction errors by a factor of three to five (\autoref{tab:response_only_metrics}), while model rankings can vary across validation windows (\autoref{sec:responseOnlyTraining}). Performance should therefore be assessed across multiple realizations and interpreted relative to the measurement-noise level, while distinguishing measurement-conditioned reconstruction from independent prediction.
    \item \emph{Assess parameter identifiability.} Before interpreting inferred physical parameters, examine their influence on the measured response and potential dependencies among their effects. Sensitivity analysis, posterior correlations, and recovery tests across parameter configurations can help distinguish genuinely identifiable quantities from parameters whose effects may be absorbed by flexible learned components.
    \item \emph{Check uncertainty before using it.} Learned noise covariances may absorb model discrepancy or become underestimated for well-fitted channels and should be compared with known sensor characteristics. Predictive interval coverage and innovation statistics should be evaluated to assess uncertainty calibration, while correlated measurement errors may require non-diagonal covariance representations. 
    \item \emph{Design for online operation from the outset.} Deployment with streaming measurements requires causal inference, favoring recursive filters or forward-only inference networks, together with computational requirements compatible with the measurement sampling rate.
\end{itemize}

\subsection{Outlook}

\paragraph{Calibrated uncertainty.}
Deep Bayesian filtering provides a probabilistic representation of latent-state inference and prediction, but the availability of posterior distributions does not in itself guarantee calibrated uncertainty estimates. Calibration has not been evaluated in the present tutorial and remains an essential requirement for applications in which uncertainty informs monitoring or operational decisions.
These distributions moreover represent only the uncertainty captured by the learned noise models, not the epistemic uncertainty of the learned networks.
Probabilistic treatments of network weights~\cite{magris2023bayesian, wang2020survey}, uncertainty quantification for learned filter components~\cite{dahan2024uncertainty, dahan2025bayesian}, and a consistent classification of the uncertainty sources in monitoring applications~\cite{Kamariotis2024} are needed before such estimates can support decisions.

\paragraph{Scalability.}
The formulations treated here keep the latent dimension small, through a reduced-order representation or by restricting the state to the monitored degrees of freedom.
Larger structures and denser sensor networks call for parametric and local \glspl{ROM}~\cite{vlachas2024parametric, vlachas2025reduced, giacomini2026surrogates} embedded within the filter, and for inference schemes whose cost does not grow cubically with the latent dimension, such as ensemble, low-rank or amortized approximations.
Sharing learned components across populations of similar structures~\cite{jian2025modal} is a related route to data efficiency.

\paragraph{Changing systems and unknown inputs.}
Structures in operation change through damage, deterioration, and environmental and operational variability~\cite{reynders2014output, avendano2017gaussian, fernandez2023bridge}, whereas the formulations treated here assume a system that is fixed, or fixed within each observation window, and an input that is measured.
Learned components that adapt online, for instance through sequentially updated local models~\cite{avendano2025sequential} or hypernetwork-based conditioning on the operating regime~\cite{vlachas2025beyond}, and the joint estimation of unknown inputs together with states and parameters, established for classical filters~\cite{impraimakis2022unscented, vettori2023adaptive, song2023joint, didyk2025towards}, would bring deep Bayesian filters closer to output-only monitoring conditions.

\paragraph{Balancing physics and learning.}
How to allocate explanatory power between prescribed and learned components remains largely open.
Priors or penalties on the magnitude of learned corrections, structural constraints such as dissipativity, and parameter-dependent operators through which physical parameters act on the dynamics they govern, rather than entering as additional inputs or latent states, are candidate mechanisms for keeping physically meaningful quantities identifiable.
Combining filtering with sparse or Bayesian model discovery~\cite{rosafalco2024ekf, fuentes2021equation} offers a route toward learned components that are themselves interpretable.

\paragraph{Validation beyond simulation.}
The demonstrations in this tutorial are based on simulated data from a single benchmark structure. Although the joint dynamic state and parameter estimation setting considers multiple parameter configurations, the dynamic state estimation experiments rely on a fixed physical system and a single training realization per framework. The reported results therefore illustrate methodological capabilities and limitations rather than establish a general ranking of the considered approaches.
Field measurements, shared benchmarks, and reporting that separates reconstruction, forecasting, state recovery and parameter recovery are needed to establish deep Bayesian filters as dependable tools for monitoring, digital twinning and decision support~\cite{thelen2022comprehensive, liang2024harnessing}.

\section*{CRediT authorship contribution statement}
\textbf{Wei Liu:} Conceptualization, Methodology, Software, Formal analysis, Investigation, Visualization, Writing -- original draft, Writing -- review \& editing.
\textbf{Konstantinos Vlachas:} Conceptualization, Methodology, Software, Data curation, Investigation, Visualization, Writing -- original draft, Writing -- review \& editing.
\textbf{Eleni Chatzi:} Conceptualization, Methodology, Formal analysis, Investigation, Supervision, Funding acquisition, Writing -- review \& editing.

\section*{Declaration of competing interest}
The authors declare that they have no known competing financial interests or personal relationships that could have appeared
to influence the work reported in this paper.

\section*{Data availability}
The template implementations of the \gls{PgDMM} and the Neural \gls{EKF} are available at \url{https://github.com/liouvill/PgDMM} and \url{https://github.com/liouvill/NeuralEKF}.

\section*{Acknowledgements}
The authors \textbf{KV, EC} gratefully acknowledge the funding from the European Commission under the Horizon Europe funding guarantee, for the project ‘TURING - Trustworthy Unified Robust Intelligent Generative Systems’ (grant agreement No: 101215032). 
Views and opinions expressed are however those of the author(s) only and do not necessarily reflect those of the European Union or HADEA.
The authors \textbf{KV, WL, EC} further acknowledge the support provided by the High Performance Computing group at ETH Zurich operating the Euler cluster that was used for all simulations.

\bibliographystyle{unsrt}
\bibliography{References}

\end{document}